%% file: ROAD-EnKF.tex
\documentclass{article}

\usepackage{titlesec}
\titleformat{\subsubsection}[runin]% runin puts it in the same paragraph
       {\normalfont\bfseries}% formatting commands to apply to the whole heading
       {\thesubsubsection}% the label and number
       {0.5em}% space between label/number and subsection title
       {}% formatting commands applied just to subsection title
       [.]% punctuation or other commands following subsection title

\usepackage[utf8]{inputenc}
\usepackage[left = 1in, right =1in,bottom =1in, top =1in]{geometry}
\usepackage{fullpage,amsfonts}
\usepackage{amsmath}
\usepackage{graphicx}
\usepackage{subcaption}
\usepackage{amsthm}
\usepackage{amssymb}
\usepackage{mathtools}
\usepackage{bbm}
\usepackage{stackrel}
\usepackage{cite}
\usepackage{bm}
\usepackage{tikz}
\usetikzlibrary{fit,positioning,arrows,automata}
\usepackage{algorithm}
\usepackage{algpseudocode}

\numberwithin{equation}{section}

\usepackage{xcolor}
\definecolor{forest}{rgb}{0,0.5,0.0}
\usepackage[colorlinks=true]{hyperref}
\hypersetup{citecolor=forest}      

\theoremstyle{plain}

\input{shared}

\title{Reduced-Order Autodifferentiable Ensemble Kalman Filters}
\author{Yuming Chen\thanks{University of Chicago, Chicago, IL (\email{ymchen@uchicago.edu}, \email{sanzalonso@uchicago.edu}, \email{willett@uchicago.edu)}}
\and Daniel Sanz-Alonso\footnotemark[1]
\and Rebecca Willett\footnotemark[1]}
\date{University of Chicago}

\begin{document}

\maketitle

\begin{abstract} 
This paper introduces a computational framework to reconstruct and forecast a partially observed state that evolves according to an unknown or expensive-to-simulate dynamical system. Our reduced-order autodifferentiable ensemble Kalman filters (ROAD-EnKFs) learn a latent low-dimensional surrogate model for the dynamics and a decoder that maps from the latent space to the state space. The learned dynamics and decoder are then used within an ensemble Kalman filter to reconstruct and forecast the state. Numerical experiments show that if the state dynamics exhibit a hidden low-dimensional structure, ROAD-EnKFs achieve higher accuracy at lower computational cost compared to existing methods. If such structure is not expressed in the latent state dynamics, ROAD-EnKFs achieve similar accuracy at lower cost, making them a promising approach for surrogate state reconstruction and forecasting. 
\end{abstract}

\section{Introduction}\label{sec:introduction}
Reconstructing and forecasting a time-evolving state given partial and noisy time-series data is a fundamental problem in science and engineering, with far-ranging applications in numerical weather forecasting, climate, econometrics, signal processing, stochastic control, and beyond. Two common challenges are the presence of \emph{model error} in the dynamics governing the evolution of the state, and the high \emph{computational cost} to simulate operational model dynamics. Model error hinders the accuracy of forecasts, while the computational cost to simulate the dynamics hinders the quantification of uncertainties in these forecasts. Both challenges can be alleviated by leveraging data to learn a surrogate model for the dynamics. Data-driven methods enable learning closure terms and unresolved scales in the dynamics, thus enhancing the forecast skill of existing models. In addition, surrogate models are inexpensive to simulate and enable using  a large number of particles  within ensemble Kalman or Monte Carlo methods for state reconstruction and forecasting, thus enhancing the uncertainty quantification.

This paper investigates a framework for state reconstruction and forecasting that relies on data-driven surrogate modeling of the dynamics in a low-dimensional latent space. Our reduced-order autodifferentiable ensemble Kalman filters (ROAD-EnKFs) leverage the EnKF algorithm 
to estimate by maximum likelihood the latent dynamics as well as a decoder from latent space to state space. The learned latent dynamics and decoder are subsequently used to reconstruct and forecast the state. Numerical experiments show that, compared to existing methods, ROAD-EnKFs achieve higher accuracy at lower computational cost provided that the state dynamics exhibit a hidden low-dimensional structure. When such structure is not expressed in the latent dynamics, ROAD-EnKFs achieve similar accuracy at lower cost, making them a promising approach for surrogate state reconstruction and forecasting.

Our work blends in an original way several techniques and insights from inverse problems, data assimilation, machine learning, and reduced-order modeling. First, if the state dynamics were known and inexpensive to simulate, a variety of filtering and smoothing algorithms from data assimilation (e.g. extended, ensemble, and unscented Kalman filters and smoothers, as well as particle filters) can be used to reconstruct and forecast the state. These algorithms often build on a Bayesian formulation, where posterior inference on the state  combines the observed data with a prior distribution defined using the model dynamics. Hence, learning a surrogate model for the dynamics can be interpreted as learning a prior regularization for state reconstruction and forecasting. Second, the task of learning the regularization can be viewed as an inverse problem: we seek to recover the state dynamics from partially and noisily observed trajectories. Data assimilation facilitates the numerical solution of this inverse problem by providing estimates of the hidden state. Third, our work leverages machine learning and reduced-order modeling to parameterize the dynamics in a low-dimensional latent space and learn a decoder from latent space to state space. In particular, we parameterize the decoder using recent ideas from discretization-invariant operator learning. Our numerical experiments demonstrate the computational advantage of co-learning an inexpensive surrogate model in latent space together with a decoder, rather than a more expensive-to-simulate dynamics in state space. 

\subsection{Related Work}
\paragraph{Ensemble Kalman Filters in Data Assimilation}
The EnKF algorithm, reviewed in \cite{houtekamer2016review, katzfuss2016understanding,roth2017ensemble,sanzstuarttaeb}, is a popular method for state reconstruction and forecasting in data assimilation, with applications in numerical weather forecasting, the geophysical sciences, and signal processing \cite{evensen1994sequential,evensen2009data,szunyogh2008local,whitaker2008ensemble}. The EnKF propagates $N$ equally-weighted particles through the dynamics, and assimilates new observations via Kalman-type updates computed with empirical moments. If the state dynamics are known, the EnKF can achieve accurate reconstruction with a small ensemble size $N$ even in applications where the state and the observations are high-dimensional, provided that the \emph{effective dimension} is moderate \cite{ghattas2022non}; EnKFs with a small ensemble size have a low computational and memory cost compared to traditional Kalman filters \cite{roth2017ensemble}. Ensemble Kalman methods are also successful solvers for inverse problems, as reviewed in \cite{chada2020iterative}.
In this paper, we employ the EnKF to approximate the data log-likelihood of surrogate models for unknown or expensive-to-simulate dynamics. The use of the EnKF for maximum likelihood estimation (MLE) was first proposed in \cite{stroud2010ensemble}, which adopted a derivative-free optimization approach; see also \cite{pulido2018stochastic}. Empirical studies on the likelihood computed with EnKFs and other data assimilation techniques can be found in \cite{carrassi2017estimating,metref2019estimating}. The application of the EnKF to approximate the data log-likelihood within pseudo-marginal Markov chain Monte Carlo methods for Bayesian parameter estimation was investigated in \cite{drovandi2021ensemble}; see also \cite{stroud2007sequential,stroud2018bayesian}. The paper \cite{chen2021auto} introduced derivative-based optimization of an EnKF approximation of the log-likelihood to perform state and parameter estimation in high-dimensional nonlinear systems. However, to the best of our knowledge, no prior work combines estimation of the log-likelihood via EnKFs with learning low-dimensional surrogate models, including both surrogate latent dynamics and a decoder from latent space to state space.

\paragraph{Blending Data Assimilation with Reduced-Order Models} Model reduction techniques have been employed in data assimilation to improve the state reconstruction accuracy in high-dimensional dynamical systems. The assimilation in the unstable subspace (AUS) method \cite{carrassi2008data,trevisan2010four,palatella2013lyapunov,sanz2015long,law2016filter} projects the dynamics onto a time-dependent subspace of the tangent space where the dynamics are unstable, and assimilates the observations therein. The unstable directions are determined by the Lyapunov vectors with nonnegative Lyapunov exponents, and can be approximated using discrete QR algorithms \cite{dieci2007lyapunov,dieci2015lyapunov}. The observations can also be projected onto the unstable directions to reduce the data dimension \cite{maclean2021particle}. We refer to \cite{benner2015survey} for a review of projection-based model reduction techniques. However, these methods rely on prior knowledge about the dynamics to identify the unstable subspaces and to construct the latent dynamics, and data assimilation is performed \emph{after} the subspaces are found. In contrast, our paper introduces a framework that uses data assimilation as a tool to build surrogate latent dynamics from data. Another approach to reduce the dimension of data assimilation problems exploits the conditionally Gaussian distribution of slow variables arising in the stochastic parameterization of a wide range of dynamical systems \cite{chen2022physics,chen2018conditional,majda2018strategies}. This conditional Gaussian structure can be exploited to obtain adequate uncertainty quantification of forecasts with a moderate sample size. A caveat, however, is that identifying the slow variables can be challenging in practice. As in our approach, these techniques often rely on machine learning to learn closure terms for the dynamics \cite{chen2022physics}. Finally, we refer to \cite{spantini2018inference} for a discussion on how the effective dimension of transport map methods for data assimilation can be reduced by exploiting the conditional independence structure of the reference-target pair. 

\paragraph{Merging Data Assimilation with Machine Learning} Recent developments in machine learning to model dynamical systems from data are reviewed in \cite{levine2021framework}. One line of work \cite{tandeo2015offline,ueno2014iterative, dreano2017estimating, pulido2018stochastic} embeds the EnKF and the ensemble Kalman smoother (EnKS) into the expectation-maximization (EM) algorithm for MLE \cite{dempster1977maximum}, with a special focus on estimation of error covariance matrices. The expectation step (E-step) is approximated by EnKF/EnKS with the Monte Carlo EM objective \cite{wei1990monte}. A subsequent line of work \cite{nguyen2019like,brajard2020combining,bocquet2020bayesian,farchi2021using,wikner2020combining} introduces training of a neural network (NN) surrogate model in the maximization step (M-step) based on the states filtered by the E-step. Unfortunately, it can be hard to achieve an accurate approximation of the E-step using EnKF/EnKS \cite{chen2021auto}. Another line of work \cite{naesseth2018variational,maddison2017filtering,le2017auto,corenflos2021differentiable} approximates the data log-likelihood with particle filters (PFs) \cite{gordon1993novel,doucet2009tutorial} and performs MLE using derivative-based optimization. However, the resampling step in PFs is not readily differentiable, and, in addition, PFs often collapse when the dimensions of the state and the observations are large \cite{agapiou2017importance,bengtsson2008curse}. Finally, techniques that leverage machine learning to obtain inexpensive \emph{analog} ensembles for data assimilation are starting to emerge \cite{yang2021machine}.

\paragraph{Data-Driven Modeling of Dynamical Systems with Machine Learning} Machine learning is also useful for dimensionality reduction in time-series modeling. As an important example, recurrent neural networks (RNNs) \cite{lipton2015critical,yu2019review} assimilate data into the time-evolving latent states using NN updates. The paper \cite{rubanova2019latent} models the latent state evolution in recurrent networks with NN-embedded differential equations \cite{chen2018neural}. Other types of NN updates to incorporate the data into latent states include gated recurrent units (GRU) \cite{de2019gru,jordan2021gated}, long short-term memory (LSTM) \cite{lechner2020longterm,harlim2021machine}, and controlled differential equations (CDEs) \cite{kidger2020neuralcde}. Another approach is to directly model the differential equation governing the state dynamics from observation data using regression. Such methods include sparse regression over a dictionary of candidate functions using $L_1$-regularization \cite{brunton2016discovering,tran2017exact,schaeffer2018extracting}.
These techniques rely on full observation of the state, and, importantly, on time-derivative data that are rarely available in practice and are challenging to approximate from noisy discrete-time observation data \cite{he2022asymptotic,he2022robust}. 
When the data are not guaranteed to lie in the same space as the underlying dynamics, an autoencoder structure can be jointly learned with the latent state dynamics \cite{champion2019data}. Different modeling techniques can be applied to learn the latent state dynamics, including sparse dictionary regression \cite{champion2019data}, recurrent networks \cite{gonzalez2018deep,maulik2021reduced}, and the Koopman operator learning \cite{lusch2018deep}. It is important to notice that, in contrast  to, e.g.,  \cite{brunton2016discovering,he2022much}, the focus of this paper is on state reconstruction and forecasting, rather than on obtaining an interpretable model for the dynamics.

\subsection{Outline and Main Contributions}

\begin{itemize}
    \item  \cref{sec:problemformulation} formalizes the problem setting and goals. We introduce a reduced-order  state-space model (SSM)  framework, where the dynamics are modeled in a low-dimensional latent space and learned jointly with a decoder from latent space to state space. 
    \item  \cref{sec:ROAD-EnKF} introduces our main algorithm, the reduced-order autodifferentiable ensemble Kalman filter (ROAD-EnKF). As part of the derivation of the algorithm, we discuss the use of EnKFs to estimate the data log-likelihood within reduced-order SSMs.
    \item  \cref{sec:implementationdetails} contains important  implementation considerations, including the design of the decoder, the use of truncated backpropagation to enhance the scalability for large windows of data, and the choice of regularization in latent space. 
    \item  \cref{sec:numerical} demonstrates the performance of our method in three examples: (i) a Lorenz 63 model embedded in a high-dimensional space, where we compare our approach to the SINDy-AE algorithm \cite{champion2019data}; (ii) Burgers equation, where we showcase that ROAD-EnKFs are able to forecast the emergence of shocks, a phenomenon not included in our training data-set; and (iii) Kuramoto-Sivashinky equation, a common test problem for filtering methods due to its chaotic behavior,  where the ROAD-EnKF framework provides a computational benefit over state-of-the-art methods with similar accuracies.
    \item \cref{sec:conclusions} closes with a summary of the paper and open questions for further research. 
\end{itemize}

\subsection*{Notation}
We denote by $t\in\{0,1,\dots\}$ a discrete-time index and by $n \in \{1,\dots,N\}$ a particle index. Time indices will be denoted with subscripts and particles with superscripts, so that $u_t^n$ represents a generic particle $n$ at time $t.$ We denote the particle dimension by $d_u.$ We denote $u_{t_0:t_1}  \triangleq\{u_t\}_{t=t_0}^{t_1}$ and $ u^{n_1:n_2} \triangleq \{u^n\}_{n=n_0}^{n_1}$. The collection $u_{t_0:t_1}^{n_0:n_1}$ is defined similarly.  The Gaussian density with mean $m$ and covariance $C$ evaluated at $u$ is denoted by $\Nc(u;m,C)$. The corresponding Gaussian distribution is denoted by $\Nc(m, C)$. 

\section{Problem Formulation}\label{sec:problemformulation}
In this section, we formalize and motivate our goals: reconstructing and forecasting a time-evolving, partially-observed state with unknown or expensive-to-simulate dynamics. An important step towards these goals is to learn a surrogate model for the dynamics. In Subsection \ref{ssec:settingmotivation} we consider an SSM framework where the state dynamics are parameterized and learned in order to reconstruct and forecast the state. Next, in Subsection \ref{ssec:rom}, we introduce a reduced-order SSM framework where the dynamics are modeled in a latent space and a decoder from latent space to state space is learned along with the latent dynamics. Our ROAD-EnKF algorithm, introduced in Section \ref{sec:ROAD-EnKF}, operates in this reduced-order SSM. 
\subsection{Setting and Motivation}\label{ssec:settingmotivation}
Consider a parameterized SSM of the form: 
\begin{alignat}{5}
    &\text{(dynamics)}&& \quad\quad u_t &&= F_\alpha (u_{t-1}) + \xi_t, \quad\quad && \xi_t \sim \Nc(0, Q_\beta), \quad\quad && 1 \le t \le T, \label{eq:ref_model_state}\\
    &\text{(observation)}&& \quad\quad y_t &&= H_t u_t + \eta_t, \quad\quad && \eta_t \sim \Nc(0, R_t), \quad\quad && 1 \le t \le T, \label{eq:ref_model_obs}\\
    &\text{(initialization)}&& \quad\quad u_0 &&\sim p_u(u_0).\label{eq:ref_model_init}
\end{alignat}
The state dynamics map $F_\alpha: \R^{d_u} \rightarrow \R^{d_u}$ and error covariance matrix $Q_\beta \in \R^{d_u \times d_u}$ depend on unknown parameter $\theta \triangleq (\alpha^\top, \beta^\top)^\top \in \R^{d_\theta}$. The observation matrices $H_t \in \R^{d_y\times d_u}$ and error covariance matrices $R_t \in \R^{d_y \times d_y}$ are assumed to be known and possibly time-varying. We further assume independence of all random variables $u_0$, $\xi_{1:T},$ and $\eta_{1:T}$.

Given observation data $y_{1:T}$ drawn from the SSM \eqref{eq:ref_model_state}-\eqref{eq:ref_model_init}, we aim to accomplish two goals:
\begin{enumerate}[wide,label={\bfseries Goal \arabic*:}]
    \item Reconstruct the states $u_{1:T}$.
    \item Forecast the states $u_{T+1:T+T_f}$ for some forecast lead time $T_f \ge 1$.
\end{enumerate}
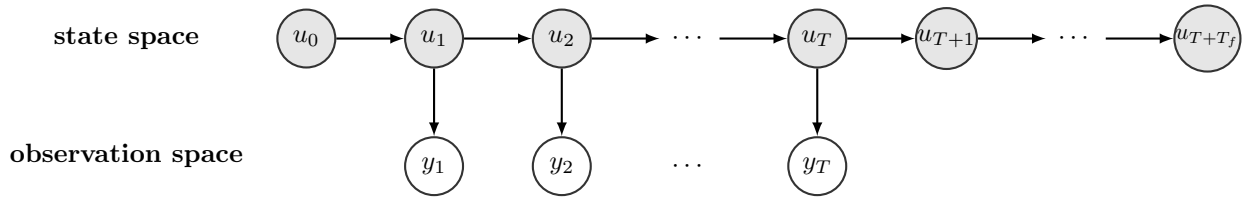
\begin{figure}[!htb]
\centering
\begin{tikzpicture}[scale = 0.5, every node/.style={scale=1}]
% \clip (-2.5,-2.5) rectangle (8.5,2.5);
\tikzstyle{main}=[circle, minimum size = 7.7mm, thick, draw =black!80, node distance = 9mm,inner sep=0pt]
\tikzstyle{connect}=[-latex, thick]
\tikzstyle{box}=[rectangle, draw=black!100]
  \node[box,draw=white!100] (State) {\textbf{state space}};
  \node[main,fill=black!10] (X0) [right=of State]{$u_0$};
  \node[main,fill=black!10] (X1) [right=of X0] {$u_1$};
  \node[main,fill=black!10] (X2) [right=of X1] {$u_2$};
  \node[main,draw=white!100] (Xdots) [right=of X2] {$\cdots$};
  \node[main,fill=black!10] (XT) [right=of Xdots] {$u_T$};
  \node[main,fill=black!10] (XT+1) [right=of XT] {$u_{T+1}$};
  \node[main,draw=white!100] (Xdots2) [right=of XT+1] {$\cdots$};
  \node[main,fill=black!10, scale=0.88] (XT+Tf) [right=of Xdots2] {$u_{T+T_f}$};
  \node[box,draw=white!100] (Observed) [below=of State] {\textbf{observation space}};
  \node[main] (Y1) [right=of Observed, below=of X1] {$y_1$};
  \node[main] (Y2) [right=of Y1,below=of X2] {$y_2$};
  \node[main,draw=white!100] (Ydots) [right=of Y2,below=of Xdots] {$\cdots$};
  \node[main] (YT) [right=of Ydots,below=of XT] {$y_T$};
  \path (X0) edge [connect] (X1);
  \path (X1) edge [connect] (X2)
        (X2) edge [connect] (Xdots)
        (Xdots) edge [connect] (XT);
  \path (XT) edge [connect] (XT+1);
  \path (XT+1) edge [connect] (Xdots2);
  \path (Xdots2) edge [connect] (XT+Tf);
  \path (X1) edge [connect] (Y1);
  \path (X2) edge [connect] (Y2);
  \path (XT) edge [connect] (YT);
\end{tikzpicture}
\caption{Structure of data under SSM \eqref{eq:ref_model_state}-\eqref{eq:ref_model_init}, where we assume only observations $y_{1:T}\triangleq \{y_1,\dots, y_T\}$ are available. Our goals are to reconstruct the states $u_{1:T}$ (Goal 1) and to forecast future states $u_{T+1:T+T_f}$ for some $T_f\ge 1$ (Goal 2).}
\label{fig:ref_data_structure}
\end{figure}
If the true parameter $\theta\in \R^{d_\theta}$ was known and the dynamics were inexpensive to simulate, the first goal can be accomplished by applying a filtering (or smoothing) algorithm on the SSM \eqref{eq:ref_model_state}-\eqref{eq:ref_model_init}, while the second goal can be accomplished by iteratively applying the dynamics model \cref{eq:ref_model_state} to the reconstructed state $u_T$. We are interested in the case where $\theta$ needs to be estimated in order to reconstruct and forecast the state.

The covariance $Q_\beta$ in the dynamics model \eqref{eq:ref_model_state} may represent model error or stochastic forcing in the dynamics; in either case, estimating $Q_\beta$ from data can improve the reconstruction and forecast of the state. In this paper, we are motivated by applications where $F_\alpha$ represents a surrogate model for the flow between observations of an autonomous ordinary differential equation (ODE). Letting $\Delta_s$ be the equally-spaced time between observations and $f_\alpha: \R^{d_u} \mapsto \R^{d_u}$ be the parameterized vector field of the differential equation, we then have 
\begin{equation}\label{eq:main_ODE_flow}
    \text{(ODE)} \quad\quad\frac{\dd u}{\dd s} = f_\alpha(u), \quad \quad F_\alpha : u(s) \mapsto u(s+\Delta_s),
\end{equation}
where $u(s)\in \R^{d_u}$ is the state as a function of continuous-time variable $s\ge 0$. The ODE \eqref{eq:main_ODE_flow} may arise from spatial discretization of a system of partial differential equations (PDEs). For instance, we will consider 1-dimensional partial differential equations for $u(x, s)$ of order $\kappa \ge 1,$ where $u$ is a function of the spatial variable $x \in [0, \mathsf{L}]$ and continuous-time variable $s \ge 0 $:
\begin{equation}\label{eq:main_PDE_flow}
    \text{(PDE)} \quad\quad\frac{\partial u}{\partial s} = f_\alpha \biggl(u, \frac{\partial u}{\partial x}, \dots,  \frac{\partial^\kappa u}{\partial x^\kappa} \biggr), \quad\quad F_\alpha: u(\cdot, s ) \mapsto u(\cdot, s + \Delta_s),
\end{equation}
with suitable boundary conditions. After discretizing this equation on a spatial domain with grid points $0 = x_1 < x_2 < \cdots < x_M = \mathsf{L} $, \cref{eq:main_PDE_flow} can be expressed in the form of \cref{eq:main_ODE_flow} by replacing the spatial derivatives with their finite difference approximations, and $u, f_\alpha, F_\alpha$ with their finite-dimensional approximations on the grid. As a result, $d_u$ equals the number of grid points $M$.  Several examples and additional details will be given in \cref{sec:numerical}. 

\subsection{Reduced-Order Modeling}\label{ssec:rom}
When the state is high-dimensional (i.e., $d_u$ is large), direct reconstruction and forecast of the state is computationally expensive, and surrogate modeling of the state dynamics map $F_\alpha$ becomes challenging. We then advocate reconstructing and forecasting the state $u_t$ through a low-dimensional latent representation $z_t$, modeling the state dynamics within the low-dimensional latent space.
This idea is formalized via the following reduced-order parameterized SSM:
\begin{alignat}{5}
    &\text{(latent dynamics)}&& \quad\quad z_t &&= \Gc_\alpha (z_{t-1}) + \zeta_t, \quad\quad && \zeta_t \sim \Nc(0, \Sc_\beta), \quad\quad && 1 \le t \le T, \label{eq:param_model_state}\\
    &\text{(decoding)}&& \quad\quad u_t &&= \Dc_{\gamma} (z_t), \quad\quad && \quad\quad && 1 \le t \le T, \label{eq:param_model_dec}\\
    &\text{(observation)}&& \quad\quad y_t &&= H_t u_t + \eta_t, \quad\quad && \eta_t \sim \Nc(0, R_t), \quad\quad && 1\le t \le T,\label{eq:param_model_obs}\\
    &\text{(latent initialization)}&& \quad\quad z_0 &&\sim p_z(z_0).\label{eq:param_model_init}
\end{alignat}
The latent dynamics map $\Gc_\alpha: \R^{d_z} \mapsto \R^{d_z}$ and error covariance matrix $\Sc_\beta \in \R^{d_z\times d_z}$ are defined on a $d_z$ dimensional latent space with $d_z < d_u$, and the decoder function $\Dc_\gamma: \R^{d_z} \mapsto \R^{d_u}$ maps from latent space to state space. The reduced-order SSM depends on an unknown parameter $\theta \triangleq (\alpha^\top, \beta^\top, \gamma^\top)^\top \in \R^{d_\theta}$. The remaining assumptions are the same as in \cref{ssec:settingmotivation}. 

\begin{figure}[!htb]
\centering
\begin{tikzpicture}[scale = 0.5, every node/.style={scale=1}]
% \clip (-2.5,-2.5) rectangle (8.5,2.5);
\tikzstyle{main}=[circle, minimum size = 7.7mm, thick, draw =black!80, node distance = 9mm,inner sep=0pt]
\tikzstyle{connect}=[-latex, thick]
\tikzstyle{box}=[rectangle, draw=black!100]
  \node[box,draw=white!100] (Latent) {\textbf{latent space}};
  \node[box,draw=white!100] (State) [below=of Latent] {\textbf{state space}};
  \node[main,fill=black!10] (Z0) [right=of Latent]{$z_0$};
  \node[main,fill=black!10] (Z1) [right=of Z0]{$z_1$};
  \node[main,fill=black!10] (Z2) [right=of Z1]{$z_2$};
  \node[main,draw=white!100] (Zdots) [right=of Z2]{$\cdots$};
  \node[main,fill=black!10] (ZT) [right=of Zdots]{$z_T$};
  \node[main,fill=black!10] (ZT+1) [right=of ZT]{$z_{T+1}$};
  \node[main,draw=white!100] (Zdots2) [right=of ZT+1] {$\cdots$};
  \node[main,fill=black!10,scale=0.88] (ZT+Tf) [right=of Zdots2] {$z_{T+T_f}$};
  \node[main,fill=black!10] (X1) [right=of State, below=of Z1] {$u_1$};
  \node[main,fill=black!10] (X2) [right=of X1, below=of Z2] {$u_2$};
  \node[main,draw=white!100] (Xdots) [right=of X2, below=of Zdots] {$\cdots$};
  \node[main,fill=black!10] (XT) [right=of Xdots, below=of ZT] {$u_T$};
  \node[main,fill=black!10] (XT+1) [right=of XT, below=of ZT+1] {$u_{T+1}$};
  \node[main,draw=white!100] (Xdots2) [right=of XT+1, below=of Zdots2] {$\cdots$};
  \node[main,fill=black!10, scale=0.88] (XT+Tf) [right=of Xdots2, below=of ZT+Tf] {$u_{T+T_f}$};
  
  \node[box,draw=white!100] (Observed) [below=of State] {\textbf{observation space}};
  \node[main] (Y1) [right=of Observed,below=of X1] {$y_1$};
  \node[main] (Y2) [right=of Y1,below=of X2] {$y_2$};
  \node[main,draw=white!100] (Ydots) [right=of Y2,below=of Xdots] {$\cdots$};
  \node[main] (YT) [right=of Ydots,below=of XT] {$y_T$};
  \path (Z0) edge [connect] (Z1);
  \path (Z1) edge [connect] (Z2)
        (Z2) edge [connect] (Zdots)
        (Zdots) edge [connect] (ZT);
  \path (ZT) edge [connect] (ZT+1);
  \path (ZT+1) edge [connect] (Zdots2);
  \path (Zdots2) edge [connect] (ZT+Tf);
  \path (Z1) edge [connect] (X1);
  \path (Z2) edge [connect] (X2);
  \path (ZT) edge [connect] (XT);
  \path (ZT+1) edge [connect] (XT+1);
  \path (ZT+Tf) edge [connect] (XT+Tf);
  \path (X1) edge [connect] (Y1);
  \path (X2) edge [connect] (Y2);
  \path (XT) edge [connect] (YT);
\end{tikzpicture}
\caption{Structure of data under reduced-order SSM \eqref{eq:param_model_state}-\eqref{eq:param_model_init}, where we assume only observations $y_{1:T}\triangleq \{y_1,\dots, y_T\}$ are available. Our goals are to reconstruct the states $u_{1:T}$ (Goal 1) and to forecast future states $u_{T+1:T+T_f}$ for some $T_f\ge 1$ (Goal 2).}
\label{fig:reduced_structure}
\end{figure}
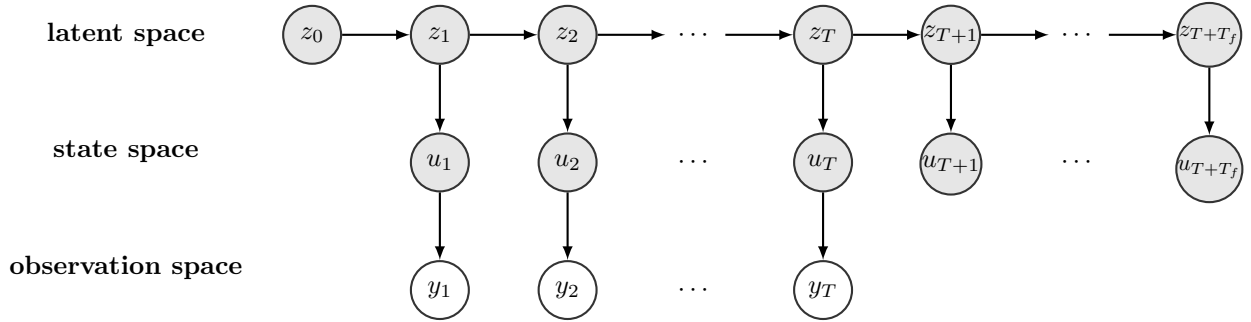

Writing $\Hc_{\gamma,t}(\cdot) \triangleq H_t \Dc_\gamma(\cdot)$, the reduced-order SSM \eqref{eq:param_model_state}-\eqref{eq:param_model_init} can be combined into
\begin{alignat}{5}
    &\text{(latent dynamics)}&&  \quad\quad z_t &&= \Gc_\alpha (z_{t-1}) + \zeta_t, \quad\quad && \zeta_t \sim \Nc(0, \Sc_\beta), \quad\quad && 1 \le t \le T, \label{eq:combined_model_state}\\
    &\text{(observation)}&&  \quad\quad y_t &&= \Hc_{\gamma,t} (z_t) + \eta_t, \quad\quad && \eta_t \sim \Nc(0, R_t), \quad\quad && 1\le t \le T,\label{eq:combined_model_obs}\\
    &\text{(latent initialization)}&& \quad\quad z_0 &&\sim p(z_0),\label{eq:combined_model_init}
\end{alignat}
where the observation function $\Hc_{\gamma,t}(\cdot)$ is nonlinear if the decoder $\Dc_\gamma(\cdot)$ is nonlinear. As in Subsection \ref{ssec:settingmotivation}, the map $G_\alpha$ may be interpreted as the flow between observations of an ODE with vector field $g_\alpha: \R^{d_z} \mapsto \R^{d_z}.$

If the true parameter $\theta \in \R^{d_\theta}$ was known, given observation data $y_{1:T}$ drawn from the reduced-order SSM \eqref{eq:param_model_state}-\eqref{eq:param_model_init}, we can reconstruct the states $u_{1:T}$ (Goal 1) by first applying a filtering (or smoothing) algorithm on \eqref{eq:combined_model_state}-\eqref{eq:combined_model_init} to estimate $z_{1:T}$, and then applying the decoder $\Dc_\gamma$. We can forecast the states $u_{T+1:T+T_f}$ (Goal 2) by first applying iteratively the latent dynamics model \cref{eq:param_model_state} to the reconstructed latent state $z_T$, and then applying the decoder $\Dc_\gamma$. As in \cref{ssec:settingmotivation}, we are interested in the case where $\theta$ needs to be estimated from the given data $y_{1:T}$.

\section{Reduced-Order Autodifferentiable Ensemble Kalman Filters}\label{sec:ROAD-EnKF}
As discussed in the previous section, to achieve both goals of state reconstruction and forecast, it is essential to obtain a suitable surrogate model for the dynamics by learning the parameter $\theta$. The general approach we take is the following: (1) estimate $\theta$ with maximum likelihood; (2) apply a filtering algorithm with estimated parameter $\theta$ to reconstruct and forecast the states $u_{1:T}$. As we shall see, the maximum likelihood estimation of $\theta$ will rely itself on a filtering algorithm.
For the SSM in \cref{ssec:settingmotivation}, this approach was introduced in \cite{chen2021auto} via AD-EnKF (Algorithm 4.1 in \cite{chen2021auto}). Here we focus on the reduced-order SSM in \cref{ssec:rom}, namely \eqref{eq:combined_model_state}-\eqref{eq:combined_model_init}, which is a more general case than the SSM in \cref{ssec:settingmotivation}; this explains the terminology reduced-order AD-EnKF (ROAD-EnKF).

In \cref{ssec:filtering}, we describe how the log-likelihood $\L(\theta)=\log p_\theta(y_{1:T})$ can be expressed in terms of the normalizing constants that arise from sequential filtering. In \cref{ssec:enkf_loglike}, we give background on EnKFs and on how to use these filtering algorithms to estimate $\L(\theta)$. In \cref{ssec:main-algorithm}, we introduce our ROAD-EnKF method that takes as input multiple independent instances of observation data $y_{1:T}^{\I}$ across the same time range, and performs both state reconstruction and forecasting.

\subsection{Sequential Filtering and Data Log-Likelihood}\label{ssec:filtering}
Suppose that $\theta= (\alpha^\top, \beta^\top, \gamma^\top)^\top$ is known. We recall that, for $1 \le t \le T,$ the \emph{filtering distributions} $p_\theta(z_t|y_{1:t})$ of the SSM \eqref{eq:combined_model_state}-\eqref{eq:combined_model_init} can be obtained sequentially, alternating between \emph{prediction} and \emph{analysis} steps:
\begin{align}
\text{(prediction)}& \quad\quad p_\theta(z_t|y_{1:t-1}) = \int \Nc\big(z_t; \Gc_\alpha(z_{t-1}), \Sc_\beta\big) p_\theta(z_{t-1}|y_{1:t-1}) \, \dd z_{t-1}, \label{eq:forecast}\\
\text{(analysis)}& \quad\quad p_\theta(z_t|y_{1:t}) = \frac{1}{\mathcal{E}_t(\theta)} \Nc\big(y_t; \Hc_{\gamma,t} (z_t), R_t\big)  p_\theta(z_t|y_{1:t-1}), \label{eq:filtering}
\end{align}
with the convention $p_\theta(\cdot|y_{1:0}) \triangleq p_\theta(\cdot)$.
% \todo{should that be $y_{1:T}$ instead of $y_{1:0}$?}
Here $\mathcal{E}_t(\theta)$ is a normalizing constant which does not depend on $z_t$. It can be shown that
\begin{equation}\label{eq:normalizing}
    \mathcal{E}_t(\theta) = p_\theta(y_t|y_{1:t-1})=\int \Nc\big(y_t; \Hc_{\gamma,t} (z_t), R_t\big)  p_\theta(z_t|y_{1:t-1}) \, \dd z_t,
\end{equation}
and therefore the data log-likelihood admits the characterization
\begin{equation}\label{eq:likecharacterization}
\L(\theta) \triangleq  \log p_\theta(y_{1:T}) = \sum_{t=1}^T \log  p_\theta(y_t|y_{1:t-1}) = \sum_{t=1}^T \log  \mathcal{E}_t(\theta).
\end{equation}
Analytical expressions of the filtering distributions $p_\theta(z_t|y_{1:t})$ and the data log-likelihood $\L(\theta)$ are only available for a small class of SSMs, which includes linear-Gaussian and discrete SSMs \cite{kalman1960new,papaspiliopoulos2014optimal}. Outside these special cases, filtering algorithms need to be employed to approximate the filtering distributions, and these algorithms can be leveraged to estimate the log-likelihood.

\subsection{Estimation of the Log-Likelihood with Ensemble Kalman Filters}\label{ssec:enkf_loglike}
Given $\theta= (\alpha^\top, \beta^\top, \gamma^\top )^\top$, the EnKF algorithm \cite{evensen1994sequential, evensen2009data} sequentially approximates  the filtering distributions $p_\theta(z_t|y_{1:t})$ using $N$ equally-weighted particles $\ton{z}.$ 
At prediction steps, each particle $\tn{z}$ is propagated using the latent dynamics model \cref{eq:combined_model_state}, while at analysis steps a Kalman-type update is performed for each particle:
\begin{alignat}{3}
\text{(prediction step)}& \quad\quad \tnf{z} = \Gc_\alpha(\tmn{z}) + \tn{\zeta}, &&\quad\quad \tn{\zeta}\iidsim \Nc(0, \Sc_\beta), \label{eq:EnKF_forecast}\\
\text{(analysis step)}& \quad\quad \tn{z} = \tnf{z} + \widehat \Kc_t \bigl(y_t + \tn{\eta} - \Hc_{\gamma,t} (\tnf{z})\bigr), &&\quad \quad \tn{\eta} \iidsim  \Nc(0, R_t ). \label{eq:EnKF_filtering}
\end{alignat}
The Kalman gain $\widehat \Kc_t \triangleq \widehat C_{zy,t} (\widehat C_{yy,t} + R_t)^{-1}$ is defined using empirical covariances given by
\begin{equation}\label{eq:empiricalcov}
    \widehat C_{zy,t} = \frac{1}{N-1} \sum_{n=1}^N (\tnf{z} - \widehat m_t) \big( \Hc_{\gamma, t}(\tnf{z}) - \widehat \Hc_t \big)^{\top}, \quad \widehat C_{yy,t} = \frac{1}{N-1} \sum_{n=1}^N \big( \Hc_{\gamma, t}(\tnf{z}) - \widehat \Hc_t \big) \big( \Hc_{\gamma, t}(\tnf{z}) - \widehat \Hc_t \big)^{\top},
\end{equation}
where
\begin{equation}\label{eq:empiricalmean}
    \widehat m_t = \frac{1}{N} \sum_{n=1}^N \tnf{z}, \quad \quad  \widehat \Hc_t = \frac{1}{N} \sum_{n=1}^N \Hc_{\gamma, t}(\tnf{z}).
\end{equation}

The empirical moments $\widehat C_{yy,t}, \widehat \Hc_t$ defined in equations \cref{eq:empiricalcov} and \cref{eq:empiricalmean} provide a Gaussian approximation to the \emph{predictive distribution} for $\Hc_{\gamma, t}(z_t)$:
\begin{equation}\label{eq:forecastdist}
    p_\theta(\Hc_{\gamma,t}(z_t)|y_{1:t-1}) \approx \Nc(\Hc_{\gamma,t}(z_t); \widehat \Hc_t, \widehat C_{yy,t}).%, \quad p_\theta(y_t|y_{1:t-1}) \approx \Nc(y_t; \widehat \Hc_t, \widehat C_{yy,t} + R_t)
\end{equation}
By applying the change of variables formula to \cref{eq:normalizing}, we have
\begin{equation}
\begin{split}
    \mathcal{E}_t(\theta) &= \int \Nc\big(y_t; \Hc_{\gamma,t}(z_t), R_t\big)  p_\theta(z_t|y_{1:t-1})\,  \dd z_t \\
    &= \int \Nc\big(y_t; \Hc_{\gamma,t}(z_t), R_t\big)  p_\theta(\Hc_{\gamma,t}(z_t)|y_{1:t-1}) \, \dd \Hc_{\gamma,t}(z_t) \\
    &\approx \Nc(y_t; \widehat \Hc_t, \widehat C_{yy,t} + R_t),
\end{split}
\end{equation}
where the approximation step follows from \cref{eq:forecastdist} and the formula for convolution of two Gaussians. From \cref{eq:likecharacterization}, we obtain the following estimate of the data log-likelihood:
\begin{equation}\label{eq:EnKF_loglike}
   \LEnKF(\theta) \triangleq \sum_{t=1}^T \log \Nc \big(y_t; \widehat \Hc_t, \widehat C_{yy,t} + R_t\big) \approx  \L(\theta).
\end{equation}
The estimate $\LEnKF(\theta)$  can be computed online with EnKF, and is stochastic as it depends on the randomness used to propagate the particles, e.g., the choice of random seed. The whole procedure is summarized in Algorithm \ref{alg:EnKF}, which implicitly defines a stochastic map $\theta \mapsto \LEnKF(\theta)$.

\begin{algorithm}
\caption{Ensemble Kalman Filter and Log-likelihood Estimation}
\label{alg:EnKF}
\begin{algorithmic}[1]
\Statex {\bf Input}:  $\theta = (\alpha^\top,\beta^\top,\gamma^\top)^\top, y_{1:T}.$ (If multiple input instances $y_{1:T}^\I$ are provided, run the following procedure for each instance $y_{1:T}^{i}$.) 
\State  {\bf Initialize} $\LEnKF(\theta)=0.$ Draw $z_0^n \iidsim  p_z(z_0)$.\label{eq:EnKF_init_alg}
\For {$t= 1, \ldots, T$}
\State Set $\tnf{z} = \Gc_\alpha(\tmn{z}) + \tn{\zeta}$, where $\tn{\zeta}\iidsim \Nc(0, \Sc_\beta)$. \label{eq:EnKF_forecast_alg} \Comment{Prediction step}
\State Compute $\widehat m_t, \widehat \Hc_t, \widehat C_{zy,t}, \widehat C_{yy,t}$ by equations \cref{eq:empiricalcov} and \cref{eq:empiricalmean} and set $\widehat \Kc_t = \widehat C_{zy,t} (\widehat C_{yy,t}+R_t)^{-1}.$ \label{eq:EnKF_mean_cov_alg}
\State  Set $\tn{z} = \tnf{z} + \widehat \Kc_t \bigl(y_t + \tn{\eta} - \Hc_{\gamma,t}(\tnf{z})\bigr)$, where $\tn{\eta}\iidsim \Nc(0, R_t)$. \label{eq:EnKF-analysis-alg} \Comment{Analysis step}
\State    Set $\LEnKF(\theta) \leftarrow \LEnKF(\theta) + \log \Nc\big( y_t; \widehat \Hc_t, \widehat C_{yy,t} + R_t \big).$  \label{eq:EnKF_like_alg}
\EndFor
\Statex {\bf Output}: EnKF particles $z_{0:T}^{1:N}$. Log-likelihood estimate $\LEnKF(\theta)$.(If multiple input instances $y_{1:T}^\I$ are provided, return instead the average of log-likelihood estimates.)
\end{algorithmic}
\end{algorithm}

\subsection{Main Algorithm}\label{ssec:main-algorithm}
The main idea of our algorithm is to perform maximum likelihood estimation on the parameter $\theta$ by gradient ascent, via differentiation through the map $\theta \mapsto \LEnKF(\theta)$.  Our core method is summarized in \cref{alg:ROAD-EnKF}, which includes estimation of $\theta$ as well as reconstruction and forecast of states. Our PyTorch implementation is at \url{https://github.com/ymchen0/ROAD-EnKF}. The gradient of the map $\theta^k \mapsto \LEnKF(\theta^k)$ can be evaluated using autodiff libraries \cite{paszke2019pytorch,jax2018github,abadi2016tensorflow} that support auto-differentiation of common matrix operations, e.g. matrix multiplication, inverse, and determinant \cite{giles2008collected}. We use the ``reparameterization trick'' \cite{kingma2013auto,rezende2014stochastic} to auto-differentiate through the stochasticity in the EnKF algorithm, as in Subsection 4.1 of \cite{chen2021auto}. 

In Section \ref{sec:numerical}, we consider numerical examples where the data are generated from an unknown SSM in the form of \cref{eq:param_model_state}-\cref{eq:param_model_init} with no explicit knowledge of the reduced-order structure; we also consider examples where the data are generated directly from  \cref{eq:ref_model_state}-\cref{eq:ref_model_init}. In practice, multiple independent instances of observation data $y_{1:T}^\I$ may be available across the same time range,  where each superscript $i\in \I$ corresponds to one instance of observation data $y_{1:T}$. We assume that each instance $y_{1:T}^i$ is drawn i.i.d. from the same SSM, with different realizations of initial state, model error, and observation error for each instance. We assume that data are split into training and test sets $y_{1:T}^{\I_\train}$ and $y_{1:T}^{\I_\test}$. During training, we randomly select a small batch of data from $y_{1:T}^{\I_\train}$ at each iteration, and evaluate the averaged log-likelihood and its gradient over the batch to perform a parameter update. The idea is reminiscent of stochastic gradient descent in the optimization literature: matrix operations of EnKF can be parallelized within a batch to utilize the data more efficiently, reducing the computational and memory cost compared to using the full training set at each iteration. The state reconstruction and forecast performance are evaluated on the unseen test set $y_{1:T}^{\I_\test}$.

% After $\theta$ is estimated, we first apply EnKF to the test observation data $y_{1:T}^{\I_\test}$ with SSM parameter $\theta$ to get filtered latent states $z_{0:T}^{\I_\test, 1:N}$. To reconstruct the states, we pass $z_{0:T}^{\I_\test, 1:N}$ through the learned decoder $\Dc_{\gamma}$ to get a particle approximation $u_{0:T}^{\I_\test, 1:N}$ of the states. To forecast the future states, we start with filtered latent states at the final iteration $z_T^{\I_\test, 1:N}$ and iteratively apply the learned latent dynamics model \cref{eq:combined_model_state} to each particle to simulate $z_{T+1:T+T_f}^{\I_\test, 1:N}$. We then pass them through the learned decoder $\Dc_{\gamma}$ to get a particle approximation $u_{T+1:T+T_f}^{\I_\test, 1:N}$ of the future states.

State reconstruction and forecast via \cref{alg:ROAD-EnKF} can be interpreted from a probabilistic point of view. For convenience, we drop the superscripts $\I$ and $k$ in this discussion. For $0\le t\le T$, since the particles $z_t^{1:N}$ form an approximation of the filtering distribution $p_\theta(z_t|y_{1:t})$ for latent state $z_t$, it follows from \cref{eq:param_model_dec} that the output particles $u_t^{1:N}$ of the algorithm form an approximation of the filtering distribution $p_\theta(u_t|y_{1:t})$ for state $u_t$. For $T+1 \le t \le T+T_f$, it follows from \cref{eq:combined_model_state} that the particles $z_t^{1:N}$ form an approximation of the predictive distribution $p_\theta(z_t|y_{1:T})$. Therefore, by \cref{eq:param_model_dec} the output particles $u_t^{1:N}$ of the algorithm form an approximation of the predictive distribution $p_\theta(u_t|y_{1:T})$ for future state $u_t$.
\begin{algorithm}
\caption{Reduced-Order Autodifferentiable Ensemble Kalman Filter (ROAD-EnKF)}
\label{alg:ROAD-EnKF}
\begin{algorithmic}[1]
\Statex {\bf Input}:  Observations $y_{1:T}^\I$, split into $y_{1:T}^{\I_\train}$ and $y_{1:T}^{\I_\test}$. Learning rate $\eta$. Batch size $B$.
\State {\bf Initialize} SSM parameter $\theta^0$ and set $k=0.$ Write $\Hc_{\gamma,t}(\cdot) = H_t \Dc_\gamma(\cdot).$
\LineComment{0em}{Training phase}
\While {not converging} %\Comment{Training}
\State Randomly select $B$ indices from $\I_\train$, denoted as $\I_B.$
\State Compute $z_{0:T}^{\I_B,1:N} , \LEnKF(\theta^k) = \textsc{EnsembleKalmanFilter}(\theta^k, y_{1:T}^{\I_B})$
 using \cref{alg:EnKF}.\label{eq:ROAD_EnKF_train_filter_alg}
\State Compute $\nt \LEnKF(\theta^k)$ by auto-differentiating the map $\theta^k \mapsto \LEnKF(\theta^k).$
\State Set $\theta^{k+1} = \theta^{k} + \eta \nt \LEnKF(\theta^k)$ and $k\leftarrow k+1.$
\EndWhile
\LineComment{0em}{Test phase}
\State $z_{0:T}^{\I_\test, 1:N},  \LEnKF(\theta^k)=\textsc{EnsembleKalmanFilter}(\theta^k, y_{1:T}^{\I_\test}).$ \label{eq:ROAD_EnKF_test_filter_alg} \Comment{State reconstruction}
\State Simulate $z_t^{\I_\test, 1:N}$ using \cref{eq:combined_model_state} with $\alpha=\alpha_k, \beta=\beta_k$ for $t = T+1, \dots, T+T_f$. \Comment{State forecast}
\State Compute $u_{0:T+T_f}^{\I_\test, 1:N} = \Dc_{\gamma_k}(z_{0:T+T_f}^{\I_\test, 1:N})$ .
%\LineComment{0em}{State Forecast} 
% \State Compute $u_{T+1:T+T_f}^{\I_\test, 1:N} = \Dc_{\gamma_k}(z_{T+1:T+T_f}^{\I_\test,  1:N})$

\Statex {\bf Output}: Learned reduced-order SSM parameter $\theta^k$ and particles $u_{0:T+T_f}^{\I_\test, 1:N}$.
\end{algorithmic}
\end{algorithm}

\section{Implementation Details}\label{sec:implementationdetails}
This section considers the practical implementation of ROAD-EnKF \cref{alg:ROAD-EnKF}, including parameterization of the surrogate latent dynamics map $g_\alpha$ and  decoder $D_\gamma$ (\cref{ssec:decoder_design}), computational efficiency for high-dimensional observations (\cref{ssec:computational_efficiency}), and regularization on latent states (\cref{ssec:latent_regularization}).

\subsection{Surrogate Latent Dynamics and Decoder Design}\label{ssec:decoder_design}
In our numerical experiments, we adopt a simple parameterization for the surrogate latent dynamics map $g_\alpha$ using a two-layer fully connected NN. For our design of the decoder $\Dc_\gamma$, the idea stems from the literature on convolutional autoencoders for computer vision tasks (e.g., \cite{mao2016image}), where both the encoder and decoder networks consist of multiple convolutional layers with residual connections that map between the image space and latent space. Here, to suit our setting, we replace the kernel-based local convolutional layers with Fourier-based spectral convolutional layers (`Fourier layers') introduced in \cite{li2020fourier,guibas2021efficient}. The latter treat a finite-dimensional vector as a spatial discretization of a function on a grid, and learn a finite-dimensional mapping that approximates an operator between function spaces. The learning accuracy is known empirically to not depend on the level of the discretization \cite{li2020fourier}, determined by $d_u$ in our case. Using Fourier layers to learn dynamical systems and differential equations was originally proposed in \cite{li2020fourier}. For the sake of completeness, we describe below the definition of spectral convolutional layers and how they are incorporated into our decoder design.
\paragraph{Spectral Convolutional Layer} Given an input $v_\text{in} \in \R^{n_\text{in} \times d_u}$ where $n_\text{in}$ is the number of input channels and $d_u$ is the input dimension, which is also the size of the grid where the function is discretized, we first apply a discrete Fourier transform (DFT) in spatial domain to get $\lambda_\text{in} \triangleq \DFT(v_\text{in}) \in \C^{n_\text{in} \times d_u}$. We then multiply it by a learned complex weight tensor $W \in \C^{n_\text{out} \times n_\text{in} \times d_u}$ that is even symmetric\footnote{That is, $W$ satisfies $W_{i,j,k} = \overline{W}_{i,j,d_u+2-k}$ $\forall i,j$ and $\forall k\ge 2$. This ensures that the inverse discrete Fourier transform of $\lambda_\text{out}$ is real. In practice, the parameterization of $W$ requires up to $n_\text{in} \times n_\text{out} \times (\lfloor d_u / 2 \rfloor + 1)$ complex entries.} to get $ \lambda_\text{out} \triangleq W \times \lambda_\text{in} \in \C^{n_\text{out} \times d_u}$. The multiplication is defined by
\begin{equation}
  (W \times \lambda_\text{in})_{i,k} =  \sum_{j=1}^{n_\text{in}} W_{i,j,k} (\lambda_\text{in})_{j,k}.
\end{equation}
This can be regarded as `channel mixing', since for the $k$-th Fourier mode ($1\le k \le d_u$), all $n_\text{in}$ input channels of $ \lambda$ are linearly mixed to produce $n_\text{out}$ output channels through the matrix $W_{\cdot, \cdot, k}$. Other types of (possibly nonlinear) mixing introduced in \cite{guibas2021efficient} can also be applied, and we leave them to future work. We then apply an inverse discrete Fourier transform (IDFT) in spatial domain to get the output $v_\text{out} = \IDFT(\lambda_\text{out}) \in \R^{n_\text{out}\times d_u}$. We call the mapping $v_\text{in} \mapsto v_\text{out}$ a spectral convolutional layer (SpecConv).

\begin{figure}[!htb]
\centering
% \begin{tikzpicture}[scale = 1, every node/.style={scale=1}]
% % \clip (-2.5,-2.5) rectangle (8.5,2.5);
% \tikzstyle{main}=[circle, minimum size = 5mm, thick, draw =black!80, node distance = 10mm]
% \tikzstyle{connect}=[-latex, thick]
% \tikzstyle{box}=[rectangle, draw=black!100]
%   \node[box] (Latent) {\textbf{Network Architecture}};
% \end{tikzpicture}
\includegraphics[width=\textwidth]{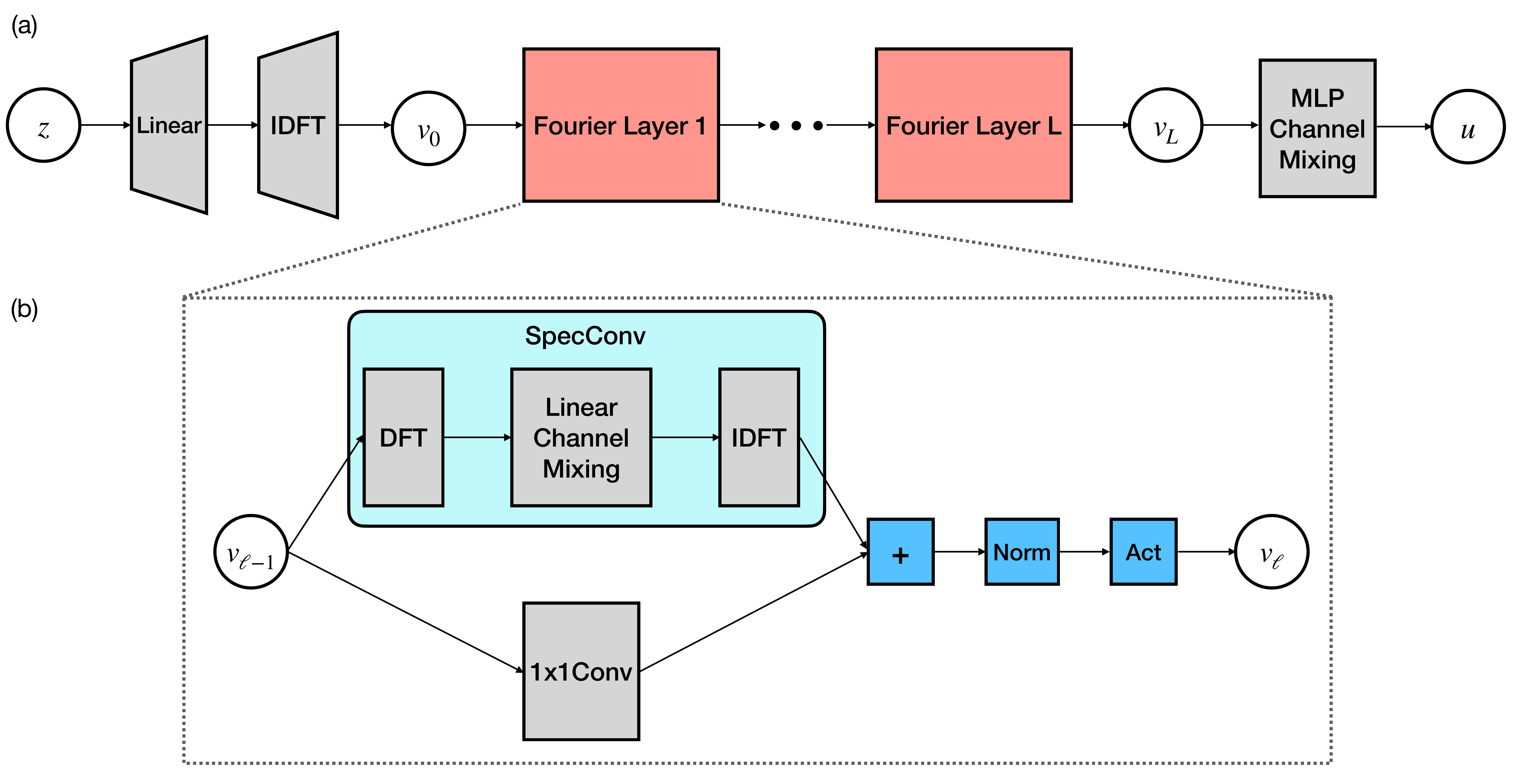}
\caption{(a) \textbf{Network architecture of the decoder $D_\gamma$.} Starting from $z\in \R^{d_z}$ in a low-dimensional latent space, we first apply a complex linear layer followed by an IDFT to lift it to $v_0 \in \R^{d_u}$ in a high-dimensional state space. We then apply $L$ Fourier layers iteratively to get $v_L \in \R^{n_L\times d_u}$ where $n_L$ is the channel dimension. We project it back to the state space by applying a two-layer fully-connected NN to mix the channels and output $u\in\R^{d_u}$. (b) \textbf{Fourier layer}: The design was first proposed in \cite{li2020fourier}, and we describe it here for the sake of completeness. The upper half represents a spectral convolutional layer, where we transform the input $v_{\ell-1}\in \R^{n_{\ell-1}\times d_u}$ into the frequency space with DFT, mix the channels with a complex linear map, and transform back with IDFT. The lower half is a one-by-one convolutional layer, which is a generalization of residual connection. The outputs from both layers are summed up and passed through a normalization and an activation layer to produce the output $v_\ell \in \R^{n_\ell \times d_u}$. }\label{fig:network}
\end{figure}

\paragraph{Fourier Neural Decoder} Given latent variable $z \in \R^{d_z}$ (where we omit the subscript $t$ for convenience), we first apply a complex linear layer $f_0(\cdot)$ to get $z_0 = f_0(z) \triangleq W_0 z + b_0\in \C^h$ for $W_0 \in \C^{h \times d_z}$ and $b_0 \in \C^h$, where $h$ is the dimension of $z_0$ to be specified. We then apply an IDFT that treats $z_0$ as a one-sided Hermitian signal in Fourier domain\footnote{$z_0$ is either truncated or zero-padded to a signal of dimension $\C^{\lfloor d_u / 2 \rfloor + 1}.$} to get $v_0 \triangleq \IDFT(z_0) \in \R^{d_u}$. We then apply $L$ spectral convolutional layers to get $v_L$, with proper choices of channel numbers as well as residual connections, normalization layers, and activation functions. More specifically, $v_L$ is defined by iteratively applying the following
\begin{equation}\label{eq:spec_layer}
    v_{\ell} = f_\ell(v_{\ell-1}):=\text{Act}\bigg(\text{Norm}\Big(\text{SpecConv}(v_{\ell-1}) + \text{1x1Conv}(v_{\ell-1}) \Big) \bigg),  \quad\quad 1\le \ell \le L,
\end{equation}
where $v_\ell \in \R^{n_\ell \times d_u}$, Act and Norm refer to the activation function and the normalization layer, 1x1Conv refers to the one-by-one convolutional layer which can be viewed as a generalization of residual connection, and $n_0 = 1$. We refer to $f_\ell$ as a `Fourier layer'. The final part of the decoder is a two-layer fully connected NN that is applied to $v_L\in \R^{n_L \times d_u}$ over channel dimension to get $u\in \R^{d_u}$. See \cref{fig:network} for the architecture. 
Notice that the learned variables $\gamma$ of the decoder include $W_0, b_0$ of the initial linear layer, complex weight tensors $W$'s of SpecConv layers, weights and biases of 1x1Conv layers, as well as the final fully-connected NN.

\subsection{Algorithmic Design for Computational Efficiency}\label{ssec:computational_efficiency}
If the time-window length $T$ is large, we follow \cite{chen2021auto} and use truncated backpropagation to auto-differentiate the map $\theta \mapsto \LEnKF(\theta)$: we divide the sequence into multiple short subsequences and backpropagate within each subsequence. The idea stems from Truncated Backpropagation Through Time (TBPTT) for RNNs  \cite{williams1995gradient, sutskever2014sequence} and the recursive maximum likelihood method for hidden Markov models \cite{le1997recursive}. By doing so, multiple gradient ascent steps can be performed for each single filtering pass, and thus the data can be utilized more efficiently. Moreover, gradient explosion/vanishing \cite{Bengio1994learning} are less likely to happen. We refer to \cite{chen2021auto} for more details. We choose this variant of ROAD-EnKF in our experiments.

In this work we are mostly interested in the case where $d_u$ and $d_y$ are large, and $d_z$ is small. Moreover, the ensemble size $N$ that we consider is moderate, i.e., $d_u \ge d_y > N  > d_z$. Therefore, we do not pursue the covariance localization approach as in \cite{chen2021auto} (see also \cite{houtekamer2001sequential,hamill2001distance}), which is most effective when $N < d_z$. Instead, we notice that the computational bottlenecks of the analysis step in the EnKF  \cref{alg:EnKF} are the $O(d_y^3)$ operations of computing the Kalman gain (\cref{eq:EnKF_mean_cov_alg}) as well as updating the data log-likelihood (\cref{eq:EnKF_like_alg}), where we need to compute the matrix inverse and log-determinant of a $d_y\times d_y$ matrix $(\widehat C_{yy,t}+R_t)$. If $d_y > N$, the number of operations can be improved to $O(N^3)$ as follows. Let $Y_t \in \R^{d_y \times N}$ be the matrix representation of the centered ensemble after applying the observation function, i.e., its $n$-th column is $Y_t^n \triangleq \frac{1}{\sqrt{N-1}}\big(\Hc_{\gamma,t}(\widehat{z}_t^n)-\frac{1}{N} \sum_{m=1}^N \Hc_{\gamma,t}(\widehat{z}_t^m) \big)$ (we drop the parameter $\gamma$ for convenience). This leads to $\widehat{C}_{yy, t} = Y_t Y_t^\top$. By the matrix inversion lemma \cite{woodbury1950inverting},
\begin{gather}
    (\widehat C_{yy,t}+R_t)^{-1} = R_t^{-1} - R_t^{-1}Y_t (I + Y_t^\top R_t^{-1} Y_t)^{-1} Y_t^\top R_t^{-1},\label{eq:inv_transform}\\
    \logdet (\widehat C_{yy,t} + R_t) = \logdet(I + Y_t^\top R_t^{-1} Y_t) + \logdet(R_t), \label{eq:loglike_transform}
\end{gather}
where $I + Y_t^\top R_t^{-1} Y_t \in \R^{N \times N}$. The computational cost can be further reduced if the quantities $R_t^{-1}$ and $\logdet(R_t)$ can be pre-computed, for instance when  $R_t=rI$ for some scalar $r\in \R$. 

Moreover, in practice, to update the ensemble in \cref{eq:EnKF-analysis-alg} of \cref{alg:EnKF}, instead of inverting $I + Y_t^\top R_t^{-1} Y_t$ directly in \cref{eq:inv_transform} followed by a matrix multiplication, we find it more numerically stable to first solve the following linear system:
\begin{equation}\label{eq:solve_system}
    (I + Y_t^\top R_t^{-1} Y_t) u_t^n = Y_t^\top R_t^{-1} \big(y_t + \gamma_t^n - \Hc_{\gamma,t}(\widehat{z}_t^n)\big) 
\end{equation}
for $u_t^n\in \R^{N}$, and then perform the analysis step (\cref{eq:EnKF-analysis-alg} of \cref{alg:EnKF}) by 
\begin{equation}
\begin{split}
    z_t^n &=  \widehat z_t^n + \widehat C_{zy, t} (\widehat C_{yy,t}+R_t)^{-1} \big(y_t + \gamma_t^n - \Hc_{\gamma,t}(\widehat{z}_t^n)\big) \\
    &= \widehat z_t^n + \widehat C_{zy, t} \big( R_t^{-1} - R_t^{-1}Y_t (I + Y_t^\top R_t^{-1} Y_t)^{-1} Y_t^\top R_t^{-1} \big)\big(y_t + \gamma_t^n - \Hc_{\gamma,t}(\widehat{z}_t^n)\big)\\
    &= \widehat z_t^n + \widehat C_{zy, t} R_t^{-1} \big(y_t + \gamma_t^n - \Hc_{\gamma,t}(\widehat{z}_t^n) - Y_t u_t^n\big).
\end{split}
\end{equation}
Similar ideas and computational cost analysis can be found in \cite{tippett2003ensemble}. For the benchmark experiments in \cref{sec:numerical}, we modify the AD-EnKF algorithm as presented in \cite{chen2021auto} to incorporate the above ideas.

\subsection{Latent Space Regularization }\label{ssec:latent_regularization}
Since the estimation of $u$ is given by $\Dc_\gamma(z),$ where both $\Dc_\gamma(\cdot)$ and $z$ need to be identified from data, we overcome potential identifiability issues by regularizing $z$ in the latent space. To further motivate the need for latent space regularization, consider the following example: if the pair $(z, \Dc_\gamma(\cdot))$ provides a good estimation of $u_t$, then so does $(c z, \frac{1}{c}\Dc_\gamma(\cdot))$ for any constant $c \ne 0$. Therefore, the norm of $z$ can be arbitrarily large, and thus we regularize $z$'s in the latent space so that their norms do not explode.
% (Regularizing the decoder: implicit or explicit?  To learn both $H$ and $z$, identifiability issues when $H$ large and $z$ small versus $z$ large and $H$ small, so we are just `fixing' one component )

We perform regularization by extending the observation model \cref{eq:combined_model_obs} to impose additional constraints on the latent state variable $z_t$'s. The idea stems from regularization in ensemble Kalman methods for inverse problems \cite{chada2020tikhonov, guth2022ensemble}. We first extend \cref{eq:combined_model_obs} to the equations:
\begin{equation}\label{eq:ext_obs_model}
\left\lbrace
\begin{array}{@{}r @{}l @{}l }
  y_t\,  &\,= \Hc_{\gamma,t}(z_t) + \eta_t, \quad\quad &\eta_t \sim \Nc(0, R_t),\\
  0\,  &\,=  z_t  + \epsilon_t, \quad\quad &\epsilon_t \sim \Nc(0, \sigma^2 I_{d_z}),
\end{array}
\right. 
\end{equation}
where $\sigma$ is a parameter to be chosen that incorporates the prior information that each coordinate of $z_t$ is an independent centered Gaussian random variable with standard deviation $\sigma$.  Define
\begin{equation}
    y_t^\aug=\begin{bmatrix}
	y_t\\ 0
    \end{bmatrix},\quad\quad
    \Hc_{\gamma,t}^\aug(z_t) = \begin{bmatrix}
	\Hc_{\gamma,t}(z_t)\\ z_t
    \end{bmatrix},\quad\quad
    \eta_t^\aug \sim \Nc(0, R_t^\aug), \quad\quad
    R_t^\aug =  \begin{bmatrix}
	R_t & 0 \\ 0 & \sigma^2 I_{d_z}
    \end{bmatrix}   .
\end{equation}
We then write \cref{eq:ext_obs_model} into an augmented observation model
\begin{equation}\label{eq:aug_obs_model}
    y_t^\aug = \Hc_{\gamma,t}^\aug(z_t) + \eta_t^\aug, \quad\quad \eta_t^\aug \sim \Nc(0, R_t^\aug).
\end{equation}
To perform latent space regularization in ROAD-EnKF, during the training stage we run EnKF (Line \ref{eq:ROAD_EnKF_train_filter_alg} of \cref{alg:ROAD-EnKF}) with augmented data $y_{1:T}^\aug$ and SSM with the augmented observation model, i.e., \eqref{eq:combined_model_state}-\eqref{eq:aug_obs_model}-\eqref{eq:combined_model_init}. During test stage, we run EnKF (Line \ref{eq:ROAD_EnKF_test_filter_alg} of \cref{alg:ROAD-EnKF}) with the original data and SSM, i.e., \eqref{eq:combined_model_state}-\eqref{eq:combined_model_obs}-\eqref{eq:combined_model_init}.

% \subsection{Improving AD-EnKF: Flexible Parameterization of State dynamics Model} \label{ssec:fno}
% In this section, we make improvements to the AD-EnKF algorithm, which is one of the baseline methods that we compare to in the experiments section. AD-EnKF is proposed in \cite{chen2021auto} that runs EnKF on the full-order SSM \eqref{eq:ref_model_state}-\eqref{eq:ref_model_init} and learns parameter $\theta = \{\alpha,\beta\}$ by auto-differentiating through a similarly defined log-likelihood objective as in \cref{ssec:enkf_loglike}. However, large dimension of $u$ imposes additional challenge on NN parameterization of $F_\alpha$ in the state dynamics model \cref{eq:ref_model_state}, and we find out empirically that the convolutional-type NN proposed in \cite{chen2021auto} does not perform well in the high-dimensional numerical experiments considered in \cref{sec:numerical}. We thus propose a more flexible NN parameterization of $F_\alpha$ using the idea of spectral convolutional layer discussed in \cref{ssec:decoder_design}.

\section{Numerical Experiments}\label{sec:numerical}
In this section, we compare our ROAD-EnKF method to the SINDy autoencoder \cite{champion2019data}, which we abbreviate as SINDy-AE. It learns an encoder-decoder pair that maps between observation space ($y_t$'s) and latent space ($z_t$'s), and simultaneously performs a sparse dictionary learning in the latent space to discover the latent dynamics. Similar to SINDy-AE, our ROAD-EnKF method jointly discovers a latent space and the dynamics therein that is a low-dimensional representation of the data. However, our method differs from SINDy-AE in four main aspects: (1) No time-derivative data for $y_{1:T}$ are required; (2) No encoder is required; (3) State reconstruction and forecast can be performed even when the data $y_{1:T}$ are noisy and partial observation of $u_{1:T}$, while SINDy-AE is targeted at noiseless and fully observed data that are dense in time; (4) Stochastic representation of latent dynamics model can be learned, and uncertainty quantification can be performed in state reconstruction and forecast tasks through the use of particles, while SINDy-AE only provides a point estimate in both tasks.

We also compare our ROAD-EnKF method to AD-EnKF \cite{chen2021auto}. Although AD-EnKF enjoys some of the benefits of ROAD-EnKF, including the capability to learn from noisy, partially observed data and perform uncertainty quantification, it directly learns the dynamics model in high-dimensional state space (i.e., on $u_t$'s instead of $z_t$'s), which leads to higher model complexity, as well as additional computational and memory costs when performing the EnKF step. Moreover, AD-EnKF does not take advantage of the possible low-dimensional representation of the state. We compare in Table \ref{tb:linear_param_est} below the capabilities of the three algorithms under different scenarios.

\begin{table}[htbp]
\centering
{\footnotesize
\begin{tabular}{|c|c|c|c|c|c|c|c|}
\hline
& \ctable{Learn from noisy}{and partially observed data} 
& \ctable{Uncertainty}{quantification}
& \ctable{No need of}{time-derivative data}
& \ctable{Low-dimensional}{state representation}
\\ \hline
{SINDy-AE\cite{champion2019data}} 
& {\xmark}
&{\xmark}
&{\xmark}
&{\cmark}
\\ \hline
{AD-EnKF\cite{chen2021auto}}
&{\cmark}
&{\cmark}
&{\cmark}
&{\xmark}
\\ \hline
{ROAD-EnKF (this paper)} 
&{\cmark}
&{\cmark}
&{\cmark}
&{\cmark}
\\ \hline
\end{tabular}}
\caption{Comparison of SINDy-AE, AD-EnKF, and ROAD-EnKF under different scenarios.}
\label{tb:linear_param_est}
\end{table}

Other alternative methods include EnKF-embedded EM algorithms (e.g. \cite{brajard2020combining}) and autodifferentiable PF algorithms (e.g., \cite{naesseth2018variational}). Since \cite{chen2021auto} already establishes AD-EnKF's superiority to those approaches, we do not include them in these experiments, and we refer to \cite{chen2021auto} for more details.

The training procedure is the following: We first specify a forecast lead time $T_f$. We then generate training data $y_{0:T}^{\I_\train}$ and test data with extended time range $(u_{0:T+T_f}^{\I_\test, \ast}, y_{0:T}^{\I_\test})$ with $N_\train \triangleq |\I_\train| $ and $N_\test \triangleq|\I_\test|$. The data are either generated from a reduced-order SSM \cref{eq:param_model_state}-\cref{eq:param_model_init} with explicit knowledge of true parameter $\theta$ (\cref{ssec:L63}), or from an SSM \cref{eq:ref_model_state}-\cref{eq:ref_model_init} with no explicit knowledge of the exact reduced-order structure (Subsections \ref{ssec:Burgers} and 
\ref{ssec:KS}). The data $y_{0:T}^{\I_\train}$ and $y_{0:T}^{\I_\test}$ are then passed into ROAD-EnKF (\cref{alg:ROAD-EnKF}), and we evaluate the following:
\paragraph{Reconstruction-RMSE (RMSE-r):} Measures the state reconstruction error of the algorithm. We take the particle mean of $u_{0:T}^{\I_\test, 1:N}$ as a point estimate of the true states $u_{0:T}^{\I_\test, \ast}$, and evaluate the RMSE:
\begin{equation}\label{eq:RMSE-r}
\rmser = \sqrt{ \frac{1}{d_u N_\test (T-T_b)} \sum_{t=T_b}^T \sum_{i\in \I_\test}\Big| \overline u_{t}^i - u_{t}^{i,\ast} \Big|^2} \,, \quad\quad \text{where } \, \overline u_{t}^i = \frac{1}{N} \sum_{n=1}^N u_{t}^{i, n}.
\end{equation}
Here $T_b$ is a number of burn-in steps to remove transient errors in the reconstruction that stem from the choice of initialization. For simplicity, we set $T_b = \lfloor T/5 \rfloor$ as in \cite{chen2021auto}.
\paragraph{Forecast-RMSE (RMSE-f):} Measures the $t$-step state forecast error of the algorithm, for lead time $t \in \{1, \ldots, T_f\}$. We take the particle mean of $u_{T+t}^{\I_\test, 1:N}$ as a point estimate of the true future states $u_{T+t}^{\I_\test, \ast}$:
\begin{equation}\label{eq:RMSE-f}
\rmsef(t) = \sqrt{ \frac{1}{d_u N_\test} \sum_{i\in \I_\test} \Big| \overline u_{T+t}^{i} - u_{T+t}^{i,\ast} \Big|^2} \,, \quad\quad \text{where } \, \overline u_{T+t}^{i} = \frac{1}{N} \sum_{n=1}^N u_{T+t}^{i, n} \, .
\end{equation}
\paragraph{Test Log-Likelihood:} Measures the averaged log-likelihood of the learned reduced-order SSM over test observation data $y_{0:T}^{\I_\test}$, which is $\LEnKF(\theta^k)$ defined in \cref{eq:ROAD_EnKF_test_filter_alg} of \cref{alg:ROAD-EnKF}.

For AD-EnKF, the above metrics can be similarly computed, following \cite{chen2021auto}. For SINDy-AE, as uncertainty quantification is not performed,  we use its decoder output as the point estimate of the state in both reconstruction and forecast. Moreover, log-likelihood computation is not available for SINDy-AE.

\subsection{Embedding of Chaotic Dynamics (Lorenz 63)}\label{ssec:L63}
In this subsection, we reconstruct and forecast a state defined by embedding a Lorenz 63 (L63) model in a high-dimensional state space. 
A similar experiment was used in \cite{champion2019data} to motivate the SINDy-AE algorithm, and hence this example provides a good point of comparison. The data are generated using the L63 system as the true latent state dynamics model:
\begin{equation}\label{eq:l63_latent}
\frac{\dd z}{\dd s} = g(z), \quad\quad
\begin{cases}
    g^{(1)}(z) = 10 (z^{(2)}-z^{(1)}),\\
    g^{(2)}(z) = z^{(1)}(28-z^{(3)})-z^{(2)},\\
    g^{(3)}(z) = z^{(1)}z^{(2)} - \frac{8}{3} z^{(3)},
\end{cases}
\quad \quad G : z(s) \mapsto z(s+\Delta_s),
\end{equation}
where $z^{(i)}$ and $g^{(i)}$ denote the $i$-th coordinate of $z$ and component of $g$, and $\Delta_s$ is the time between observations. We further assume there is no noise in the true latent state dynamics model, i.e., $S = 0$. To construct the true reduced-order SSM, we define $D \in \R^{d_u \times 6}$ such that its $i$-th column $D^i\in \R^{d_u}$ is given by the discretized $i$-th Legendre polynomial over $d_u$ grid points. The true states $u_t\in \R^{d_u}$ are defined by
\begin{equation}\label{eq:l63_dec}
u_t\triangleq D
\begin{bmatrix}
z_t^{(1)}/40 &
z_t^{(2)}/40 &
z_t^{(3)}/40 &
(z_t^{(1)}/40)^3 &
(z_t^{(2)}/40)^3 &
(z_t^{(3)}/40)^3 &
\end{bmatrix}^\top.
\end{equation}
We consider two cases of the observation model \eqref{eq:param_model_obs}: (1) full observation, where all coordinates of $u_t$ are observed, i.e.,  $H_t=I_{d_u}$ and $d_y=d_u$; (2) partial observation, where for each $t$, only a fixed portion $c<1$ of all coordinates of $u_t$ are observed, and the coordinate indices are chosen randomly without replacement. In this case, $H_t\in \R^{d_y\times d_u}$ is a submatrix of $I_{d_u}$ and varies across time, and $d_y = c d_u$. This partial observation set-up has been studied in the literature (e.g., \cite{brajard2020combining,bocquet2020bayesian}) for data assimilation problems. For both cases, we assume $R_t = 0.01 I_{d_y}$ and $z_0 \sim \Nc(0, 4I_{d_z})$. 

We consider full observation with $d_u=d_y=128$ and partial observation  with $d_u=128$, $d_y=64$ (i.e., $c=1/2$). We generate $N_\train=1024$ training data and $N_\test=20$ test data with the true reduced-order SSM defined by \eqref{eq:l63_latent} and \eqref{eq:l63_dec}. We set the number of observations $T=250$ with time between observations $\Delta_s=0.1$. We set the forecast lead time $T_f=10$. The latent flow map $G$ is integrated using the Runge–Kutta–Fehlberg method. The surrogate latent dynamics map $g_\alpha$ is parameterized as a two-layer fully connected NN, and is integrated using a fourth-order Runge-Kutta method with step size $\Delta_s^\text{int}=0.05$. The error covariance matrix $S_\beta$ in the latent dynamics is parametrized using a diagonal matrix with positive diagonal elements $\beta\in \R^{d_z}$. The decoder $\Dc_\gamma$ is parameterized as a Fourier Neural Decoder (FND) discussed in \cref{ssec:decoder_design}. Details of the network hyperparameters for this and subsequent examples are summarized in \cref{tb:exp_network_param}, obtained through cross-validation experiments on the training dataset. The latent space dimension for both SINDy-AE and ROAD-EnKF is set to  $d_z=3$. The ensemble size for both AD-EnKF and ROAD-EnKF is set to $N=100$.
\begin{table}[htbp]
\centering
\begin{tabular}{|c|c|c|c|c|}
\hline
& & L63 & Burgers & KS \\ \hline
\multirow{5}{*}{FND} & $L$ & 4 & 2 & 4\\ \cline{2-5}
                     & $h$ & 6 & 40 & 40\\ \cline{2-5}
                     & $(n_0,\dots, n_L)$ & (1, 20, 20, 20, 20) & (1, 20, 20) & (1, 20, 20, 20, 20) \\ \cline{2-5}
                     & Norm & \multicolumn{3}{c|}{LayerNorm} \\ \cline{2-5}
                     & Activation & \multicolumn{3}{c|}{ReLU}\\ \hline
Latent space reg.    & $\sigma$  & 2 & 4 & 4\\ \hline
\multirow{4}{*}{Optimization}& Optimizer & \multicolumn{3}{c|}{Adam}\\ \cline{2-5}
                             & Learning rate ($\eta$) & \multicolumn{3}{c|}{1e-3}\\ \cline{2-5}
                             & Batch size ($B$) & 16 & 4 & 4\\ \cline{2-5}
                             & TBPTT length & \multicolumn{3}{c|}{10}\\ \hline
\end{tabular}
\caption{Choices of hyperparameters for ROAD-EnKF on different numerical examples.}
\label{tb:exp_network_param}
\end{table}

\begin{figure}[!htb]
\centering
\includegraphics[width=\textwidth]{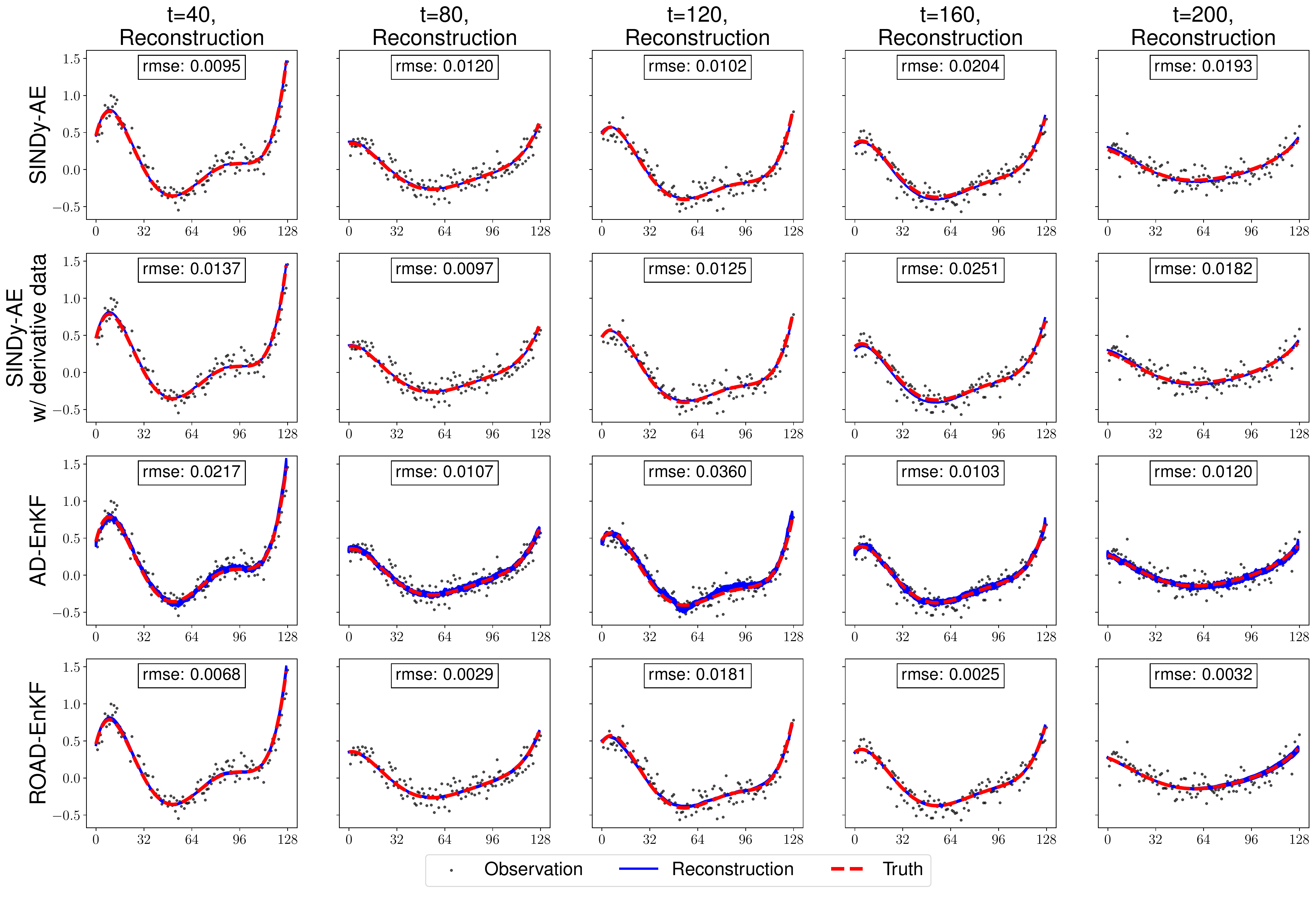}
\caption{State reconstruction performance with full observation ($d_u=d_y=128$) on the embedded L63 example in \cref{ssec:L63}. For each method (row),  the reconstructed states $u_t$ (blue) for a single test sequence are plotted for $t=40, 80, 120, 160, 200$ (column). The true values of the 128-dimensional states are plotted in red dashed lines, along with the noisy observations in black dots. Both AD-EnKF and ROAD-EnKF perform probabilistic state reconstructions through particles (all plotted in blue), while SINDy-AE only provides point estimates. The reconstruction RMSE's are computed for each plot. For SINDy-AE, even \emph{with} derivative data (not required for AD-EnKF and ROAD-EnKF), the reconstruction performance is similar to that of AD-EnKF, while being worse than that of ROAD-EnKF.  }\label{fig:l63_recon}
\end{figure}

\begin{figure}[!htb]
\centering
\includegraphics[width=\textwidth]{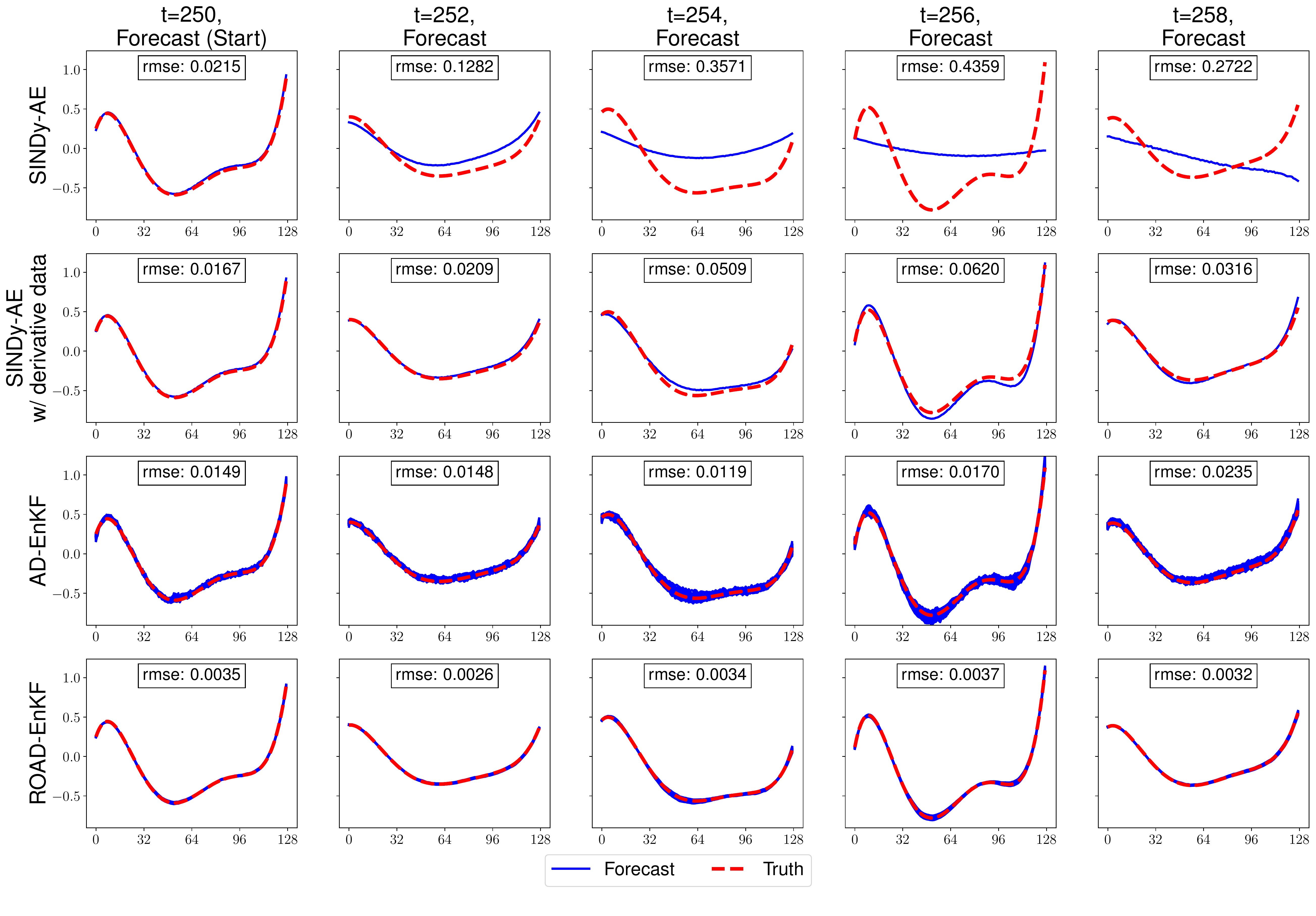}
\caption{Forecast performance with full observation ($d_u=d_y=128$) on the embedded L63 example in \cref{ssec:L63}. For each method (row),  the forecasted states $u_t$ (blue) for a single test sequence are plotted for $t=250$ (start of forecast), $252, 254, 256, 258$ (column). The true values of the $d_u=128$ dimensional states are plotted in red dashed lines. Both AD-EnKF and ROAD-EnKF perform probabilistic forecast through particles (all plotted in blue), while SINDy-AE only provides point estimates. The forecast RMSE's are computed for each plot. For SINDy-AE, even \emph{with} derivative data (not required for AD-EnKF and ROAD-EnKF), the forecast performance is similar to that of AD-EnKF, while being worse than that of ROAD-EnKF. }\label{fig:l63_pred}
\end{figure}

\begin{figure}[!htb]
\centering
\begin{subfigure}[b]{\textwidth}
\includegraphics[width=\textwidth]{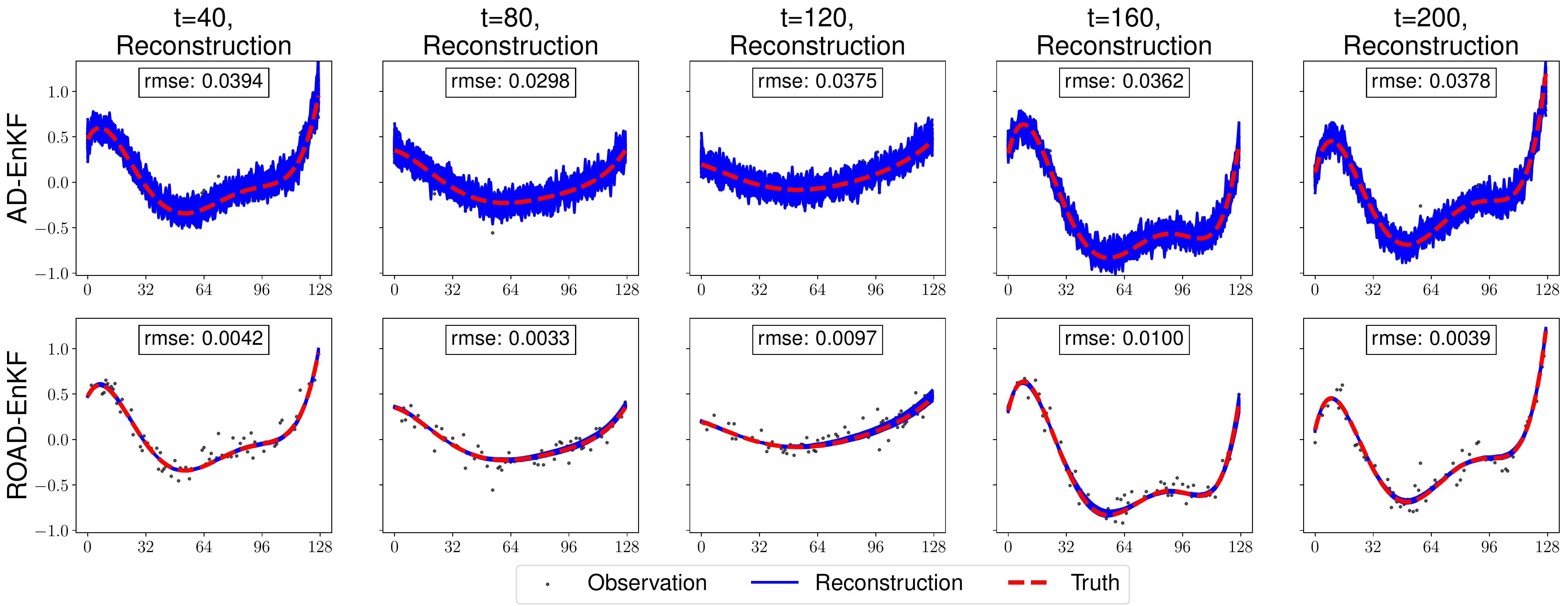}
\end{subfigure}
\begin{subfigure}[b]{\textwidth}
\includegraphics[width=\textwidth]{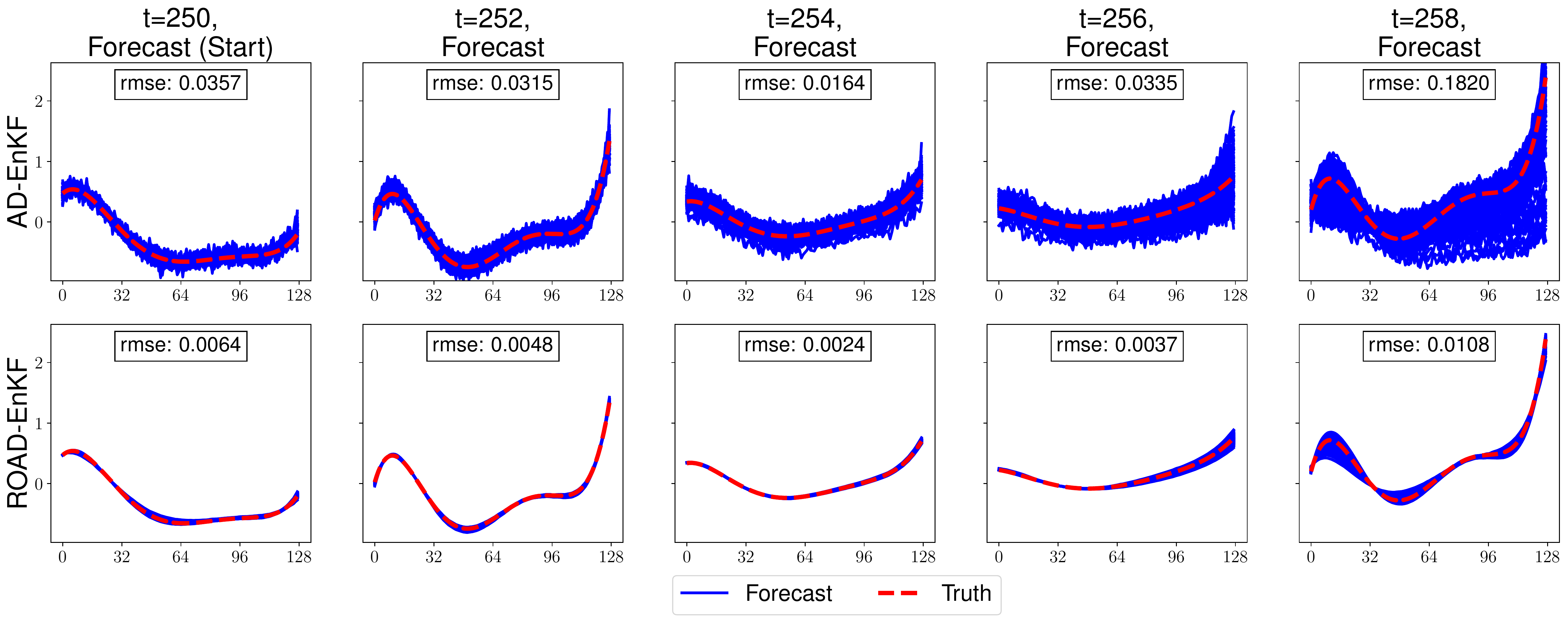}
\end{subfigure}
\caption{State reconstruction (upper half) and forecast (lower half) performance with partial observation ($d_u=128$, $d_y=64$) on the embedded L63 example in \cref{ssec:L63}. For each method, the reconstructed states $u_t$ (blue) for a single test sequence are plotted for $t=40, 80, 120, 160, 200$ (column), and the forecasted states $u_t$ (blue) for a single test sequence are plotted for $t=250$ (start of forecast), $252, 254, 256, 258$ (column). The true values of the 128-dimensional states are plotted in red dashed lines, along with the noisy observations in black dots. SINDy-AE is inapplicable here because it cannot handle partial observations, while both AD-EnKF and ROAD-EnKF perform probabilistic state reconstructions and forecast through particles (all plotted in blue). The reconstruction/forecast RMSEs are computed for each plot.}\label{fig:l63_partial_recon}
\end{figure}

In \cref{tb:l63_table} we list the performance metrics of each method with full and partial observation. The state reconstruction and forecast performance on a single instance of test data are plotted in  \cref{fig:l63_recon,fig:l63_pred} for the full observation case, and in \cref{fig:l63_partial_recon} for the partial observation case. For the full observation case, we compare ROAD-EnKF with AD-EnKF and SINDy-AE, adopting for the latter the implementation in  \cite{champion2019data}. Since SINDy-AE requires time-derivative data as input, we use a finite difference approximation computed from data $y_{1:T}$. We also include the results for SINDy-AE where the exact time-derivative data are used. We find that ROAD-EnKF is able to reconstruct and forecast the states consistently with the lowest RMSE, and the performance is not affected by whether the state is fully or partially observed. AD-EnKF is able to reconstruct and forecast the state with a higher RMSE than that of ROAD-EnKF, and the performance deteriorates in the partially observed setting. SINDy-AE with finite difference approximation of derivative data also achieves higher reconstruction RMSE than that of ROAD-EnKF, and does not give accurate state forecasts. This is likely due to the fact that data are sparse in time (i.e., $\Delta_s$ is large) which leads to a larger error when approximating the true time-derivative, and hence it is more difficult to extract meaningful dynamics from the data. Even when the true time-derivative data are used (which is not available unless we have explicit knowledge of the true reduced-order SSM), SINDy-AE has a higher reconstruction RMSE compared to ROAD-EnKF, and its forecast performance is still worse than the other two methods. Moreover, it cannot handle partial observation.

In terms of computational cost, ROAD-EnKF is more efficient than AD-EnKF since the surrogate dynamics are cheaper to simulate and the EnKF algorithm is more efficient to perform in both training and testing. However, ROAD-EnKF takes more time than SINDy-AE, since the latter does not rely on a filtering algorithm, but rather an encoder, to reconstruct the states and perform learning.

% \begin{figure}[!htb]
% \begin{subfigure}[c]{\linewidth}
% \centering
% \includegraphics[height=5cm]{L63/new_bar.pdf}
% % \includegraphics[width=0.6\textwidth]{L63/bar.pdf}
% \subcaption{Comparison of RMSE-r, RMSE-f(1), and RMSE-f(5).}
% \vspace{0.5cm}
% \end{subfigure}
% \begin{subtable}{\textwidth}
\begin{table}[!htb]
\centering
{\footnotesize
\begin{tabular}{|c|c|c|c|c?{0.5mm}c|c|}
\hline
          & \ctable{SINDy-AE}{(full)} & \ctable{SINDy-AE}{(w/ derivative, full)} & \ctable{AD-EnKF}{(full)} & \ctable{ROAD-EnKF}{(full)} & \ctable{AD-EnKF}{(partial)} & \ctable{ROAD-EnKF}{(partial)}  \\ \hline
RMSE-r    &  0.0142 & 0.0148  & 0.0168  & 0.0078 & 0.0368 & 0.0079    \\ \hline
RMSE-f(1) &  0.1310 & 0.0191  & 0.0156 & 0.0069 & 0.0315 & 0.0069    \\ \hline
RMSE-f(5) &  1.6580 & 0.0333  & 0.0335 & 0.0141 & 0.0729 & 0.0125    \\ \hline
Log-likelihood &  \multicolumn{2}{c|}{$-$}    & $2.25\times 10^4$ & $2.58\times 10^4$ & $1.28\times 10^4$ & $1.40\times 10^4$    \\ \hline
Training time (per epoch)&    \multicolumn{2}{c|}{5.15s}  & 9.74s  & 6.15s & 8.86s & 5.62s    \\ \hline
Test time&  \multicolumn{2}{c|}{2.35s} & 4.57s & 2.95s & 4.52s  & 2.73s   \\ \hline
\end{tabular}}
% \subcaption{Comparison of log-likelihood, training time, and test time. Log-likelihood evaluation is not supported for SINDy-AE.}
% \end{subtable}
\caption{Performance metrics for different algorithms at convergence. (Embedded L63 example, \cref{ssec:L63}.)}\label{tb:l63_table}
\end{table}

% \begin{table}[!htb]
% \centering
% \begin{tabular}{|c|c|c|}
% \hline
%           & AD-EnKF & ROAD-EnKF  \\ \hline
% RMSE-r     & 0.0368  & 0.0079     \\ \hline
% RMSE-f(1)  & 0.0315  & 0.0069     \\ \hline
% RMSE-f(5)  & 0.0729  & 0.0125     \\ \hline
% Log-likelihood & $1.28\times 10^4$ & $1.40\times 10^4$    \\ \hline
% % Training time (per epoch)& 9.51s  & 5.85s     \\ \hline
% % Test time (per epoch)& 4.05s  & 2.57s     \\ \hline
% \end{tabular}
% \caption{Table of performance metrics for different algorithms at convergence. (Embedded L63 example, partial observation, \cref{ssec:L63})}
% \label{tb:l63_table}
% \end{table}

\subsection{Burgers Equation}\label{ssec:Burgers}
In this subsection and the following one, we learn high-dimensional SSMs without explicit reference to a true model for low-dimensional latent dynamics. We first consider the 1-dimensional Burgers equation for $u(x, s)$, where $u$ is a function of the spatial variable $x \in [0,\mathsf{L}]$ and continuous-time variable $s>0$:
\begin{equation}\label{eq:burgers}
\begin{split}
    &\frac{\partial u}{\partial s} = -u \frac{\partial u}{\partial x} + \nu \frac{\partial^2 u}{\partial x^2},\\
    &u(0, s) = u(\mathsf{L}, s) = 0,\\
    &u(x, 0) = u_0(x).
\end{split}
\end{equation}
Here $\nu$ is the viscosity parameter, and we set $\nu = 1/150$, $\mathsf{L}=2$. Burgers equation \cite{burgers1948mathematical} has various applications in fluid dynamics, including modeling of viscous flows. We are interested in reconstructing solution states, as well as in the challenging problem of forecasting shocks that emerge outside the time range covered by the training data. Equation \cref{eq:burgers} is discretized on $[0,\mathsf{L}]$ with equally-spaced grid points $0=x_1<x_2<\cdots<x_M=\mathsf{L}$, using a second-order finite difference method. Setting $\Delta x:= x_i - x_{i-1} = \frac{\mathsf{L}}{M-1}$, we obtain the following ODE system:
\begin{equation}\label{eq:burgers_discretized}
\begin{split}
&\frac{\dd u^{(i)}}{\dd s} = - \frac{\big( u^{(i+1)} \big)^2-\big( u^{(i-1)} \big)^2}{4\Delta x} + \nu \frac{u^{(i+1)}-2u^{(i)}+u^{(i-1)}}{\Delta x^2}, \quad\quad i=2,\dots, M-1,\\
&u^{(1)}(s) = u^{(M)}(s) = 0, \\
&u^{(i)}(0) = u_0(i\Delta x).
\end{split}
\end{equation}
Here $u^{(i)}(s)$ is an approximation of $u(i\Delta x, s)$, the value of $u$ at the $i$-th spatial node at time $s$. Equation \Cref{eq:burgers_discretized} defines a flow map $F:u(s) \mapsto u(s+\Delta_s)$ for state variable $u$ with $d_u=M$, which we refer to as the true state dynamics model. We assume there is no noise in the dynamics, i.e., $Q=0$.

Similar to \cref{ssec:L63}, we consider two cases: full observation with $d_u = d_y = 256$ and partial observation with $d_u=256$, $d_y=128$ (i.e., $c=1/2$). The initial conditions $u_0$ are generated in the following way:
\begin{equation}\label{eq:burgers_discretized_init}
    u_0^{(i)} = U \sin \frac{2\pi i \Delta x}{\mathsf{L}}, \quad\quad U\sim \text{Uniform}(0.5, 1.5).
\end{equation}
We generate $N_\train=1024$ training data and $N_\test=20$ test data with the true state dynamics model defined through \cref{eq:burgers_discretized,eq:burgers_discretized_init} with $R_t = 0.01 I_{d_y}$. We set the number of observations $T=300$ with time between observations $\Delta_s=0.001$. We set the forecast lead time $T_f=300$. The flow map $F$ is integrated using the fourth-order Runge–Kutta method with a fine step size $\Delta_s/20$. The surrogate latent dynamics map $g_\alpha$ is parameterized as a two-layer fully connected NN, and is integrated using a fourth-order Runge-Kutta method with step size $\Delta_s^\text{int} = 0.001$. The error covariance matrix $S_\beta$ in the latent dynamics is parametrized using a diagonal matrix with positive diagonal elements $\beta\in \R^{d_z}$. The decoder $\Dc_\gamma$ is parameterized as an FND, discussed in \cref{ssec:decoder_design}. Details of the network hyperparameters are listed in \cref{tb:exp_network_param}. The latent space dimension for ROAD-EnKF is set to $d_z=40$. The ensemble size for both AD-EnKF and ROAD-EnKF is set to $N=100$. In this example and the following one, we set $z_0\sim \Nc(0, \sigma^2 \I_{d_z})$ with the same $\sigma$ defined in \cref{ssec:latent_regularization}.

\begin{figure}[H]
\centering
\begin{subfigure}[b]{\textwidth}
\includegraphics[width=\textwidth]{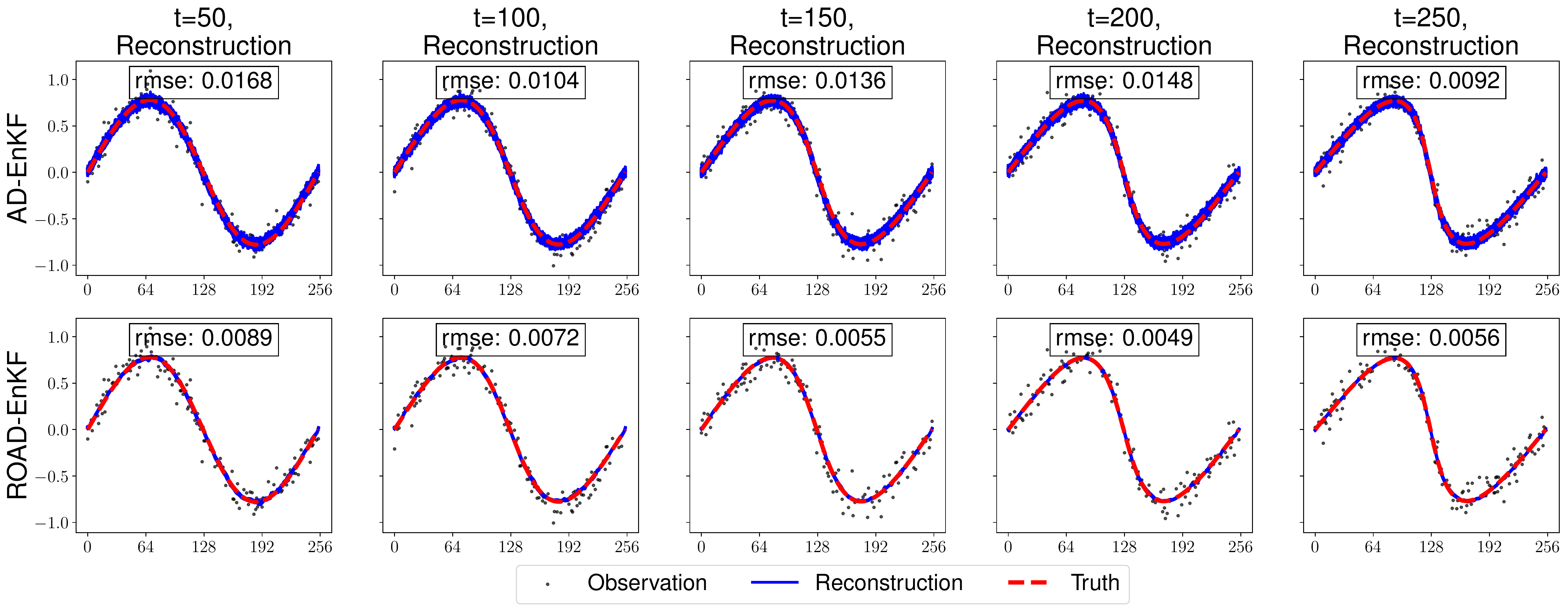}
\end{subfigure}
\begin{subfigure}[b]{\textwidth}
\includegraphics[width=\textwidth]{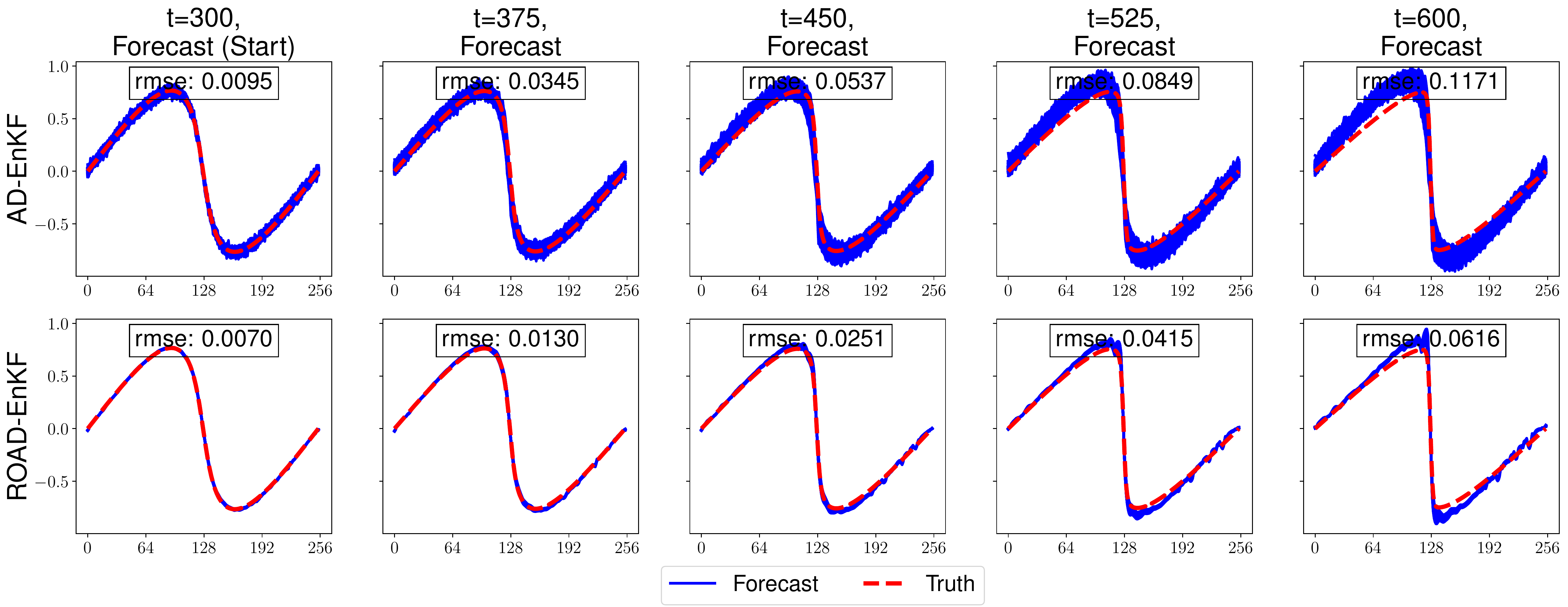}
\end{subfigure}
\caption{State reconstruction (upper half) and forecast (lower half) performance with partial observation ($d_u=256$, $d_y=128$) on the Burgers example in \cref{ssec:Burgers}. For each method, the reconstructed states $u_t$ (blue) for a single test sequence are plotted for $t=50, 100, 150, 200, 250$ (column), and the forecasted states (blue) for a single test sequence are plotted for $t=300$ (start of forecast), $375, 450, 525, 600$ (column). The true values of the 256-dimensional states are plotted in red dashed lines, along with the noisy observations in black dots. Both AD-EnKF and ROAD-EnKF perform probabilistic state reconstructions and forecast through particles (all plotted in blue). The reconstruction/forecast RMSEs are computed for each plot.  }\label{fig:burgers_recon_partial}
\end{figure}

\begin{figure}[H]
\centering
\begin{subfigure}[b]{0.55\textwidth}
   \includegraphics[width=1\linewidth]{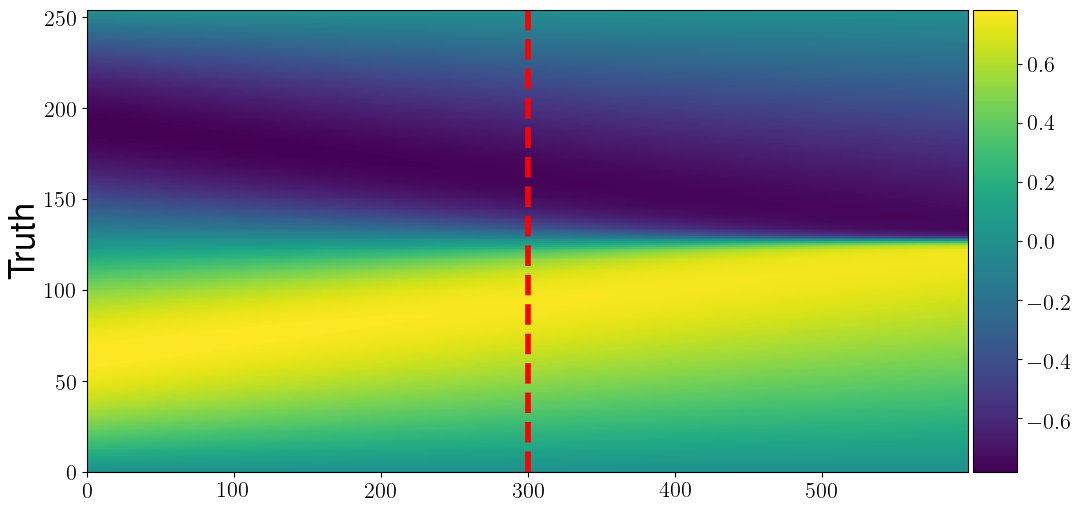}
   \caption{Ground truth.}
\end{subfigure}

\begin{subfigure}[b]{\textwidth}
   \includegraphics[width=1\linewidth]{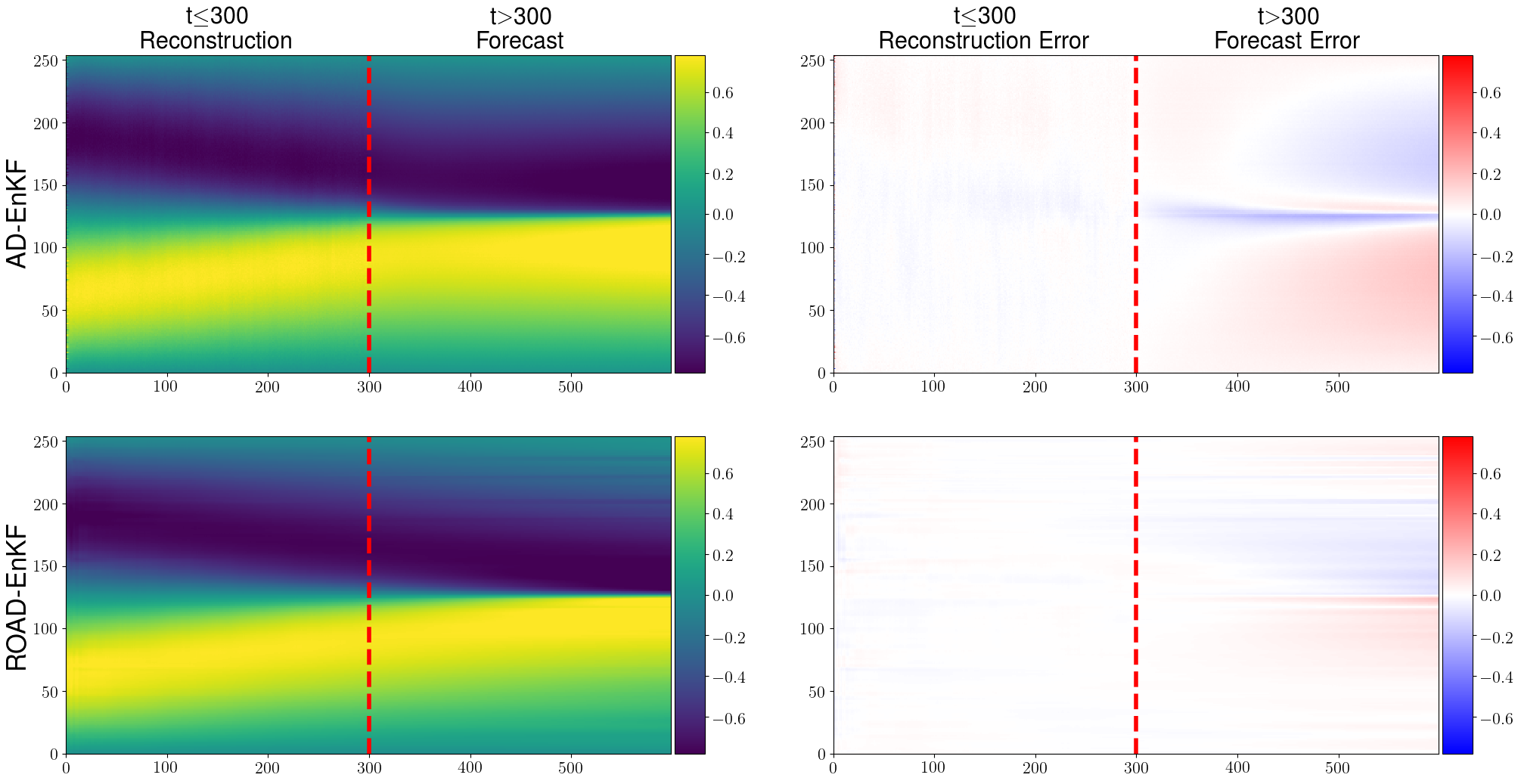}
   \caption{Reconstruction and forecast.}

\end{subfigure}
\caption{Contour plot of state reconstruction and forecast output with partial observation ($d_u=256$, $d_y=128$) on the Burgers example in \cref{ssec:Burgers}, as well as the ground truth (top). For each method (row), the reconstructed and forecasted states (left column) for a single test sequence are plotted, for each state dimension (y-axis) and time (x-axis). The error compared to the ground truth are plotted in the right column. For both AD-EnKF and ROAD-EnKF we use particle means as point estimates. }
\label{fig:burgers_contour_partial} 
\end{figure}

In \cref{tb:burgers_table}, we list the performance metrics of each method with full and partial observation. The state reconstruction and forecast performance on a single instance of test data are plotted in Figures \ref{fig:burgers_recon_partial} (snapshots) and \ref{fig:burgers_contour_partial} (contour plot) for the partial observation case. Corresponding plots with full observation are shown in Figures \ref{fig:burgers_recon} and \ref{fig:burgers_contour} in the appendix. We find that ROAD-EnKF is able to reconstruct and forecast the states with the lowest RMSE, in both full and partial observation scenarios. More importantly, the emergence of shocks is accurately forecasted even though this phenomenon is not included in the time range covered by the training data. AD-EnKF achieves a higher RMSE than ROAD-EnKF for both state reconstruction and forecast tasks. 
AD-EnKF forecasts the emergence of shocks with lower accuracy than ROAD-EnKF, which indicates that AD-EnKF fails to fully learn the state dynamics. SINDy-AE with finite difference approximation of derivative data has the highest reconstruction RMSE among the three methods, and is not able to produce meaningful long-time state forecasts. This is remarkable, given that in this example the data are relatively dense ($\Delta_s$ is small) which facilitates, in principle, the approximation of time derivatives. In terms of computational cost, ROAD-EnKF is more efficient than AD-EnKF during both training and testing, but takes more time than SINDy-AE for the same reason as in \cref{ssec:L63}. 

In \cref{tb:burgers_table_2}, we list the performance metrics of ROAD-EnKF with full observation and different choices of latent space dimension $d_z$ ranging from 1 to 240. The results for partial observation show a similar trend and are not shown. We find that, as $d_z$ increases, the state reconstruction performance stabilizes when $d_z \ge 4$. In order to achieve better long-time state forecast performance, $d_z$ needs to be further increased, and the forecast performance stabilizes when $d_z \ge 10$. Both training and testing time slightly increase as $d_z$ grows, which can be explained by the following: The computational time for both training and testing can be divided into the prediction step and the analysis step. We have shown in \cref{ssec:computational_efficiency} that the computational bottleneck of the analysis step depends on the choices of ensemble size $N$ and $d_y$, and is less affected by the increase of $d_z$. Moreover, the computational time of the prediction step depends on the complexity of the surrogate latent dynamics (two-layer NNs), which are relatively cheap to simulate for ROAD-EnKF. On the other hand, AD-EnKF enjoys similar computational complexity as ROAD-EnKF during the analysis step, but requires a more complicated surrogate model (NNs with Fourier layers) to capture the dynamics, which is more expensive to simulate. More experimental results on different parameterization methods of surrogate dynamics can be found in \cref{tb:burgers_euler_table} in the appendix. 

% \begin{figure}[!htb]
% \begin{subfigure}[c]{\linewidth}
% \centering
% \includegraphics[height=5cm]{Burgers/new_bar_1.pdf}
% \subcaption{Comparison of RMSE-r, RMSE-f(30), and RMSE-f(150).}
% \vspace{0.5cm}
% \end{subfigure}
% \begin{subtable}{\textwidth}
\begin{table}[!htb]
\centering
\begin{tabular}{|c|c|c|c?{0.5mm}c|c|}
\hline
          &\ctable{SINDy-AE}{(full)} & \ctable{AD-EnKF}{(full)} & \ctable{ROAD-EnKF}{(full)} & \ctable{AD-EnKF}{(partial)} & \ctable{ROAD-EnKF}{(partial)}  \\ \hline
RMSE-r    &  0.1433 & 0.0102 & 0.0044 & 0.0122 &0.0081     \\ \hline
RMSE-f(30) &  4.4579 & 0.0212  & 0.0096 & 0.0228 & 0.0160    \\ \hline
RMSE-f(150) &  4.4906 & 0.0763  & 0.0514 & 0.0724 & 0.0581     \\ \hline
Log-likelihood &  $-$& $6.40\times 10^4$ & $6.60\times 10^4$  &$3.24 \times 10^4$ & $3.27 \times 10^4$   \\ \hline
Training time (per epoch)&  11.78s  & 26.75s & 12.10s & 27.08s & 12.20s    \\ \hline
Test time&  2.78s & 11.54s & 4.21s &7.76s & 3.24s    \\ \hline
\end{tabular}
% \subcaption{Comparison of log-likelihood, training time, and test time. Log-likelihood evaluation is not supported for SINDy-AE.}
% \end{subtable}
\caption{Performance metrics for different algorithms at convergence. (Burgers example, \cref{ssec:Burgers}.)}\label{tb:burgers_table}
\end{table}

%\begin{figure}[!htb]
%\begin{subfigure}[c]{\linewidth}
%\centering
%\includegraphics[height=6cm]{Burgers/new_bar_2.pdf}
% \vspace{-0.6cm}
%\subcaption{Comparison of RMSE-r, RMSE-f(30), and RMSE-f(150).}
%\vspace{0.5cm}
%\end{subfigure}
\begin{table}[!htb]
\centering
{\footnotesize
\begin{tabular}{|c|c|c|c|c|c|c|c|c|c|}
\hline
&\multicolumn{8}{c|}{ROAD-EnKF} & \multirow{2}{*}{AD-EnKF}\\
\cline{2-9}
          &$d_z=1$ & $d_z=2$ & $d_z=4$ & $d_z=10$ & $d_z=20$ & $d_z=40$ & $d_z=120$ & $d_z=240$ &   \\ \hline
 RMSE-r    &  0.2293 & 0.0316 & 0.0035 & 0.0058 & 0.0039 & 0.0044 & 0.0048 & 0.0059 & 0.0102     \\ \hline
 RMSE-f(30) &  0.2593 & 0.0761 & 0.0165  & 0.0108 & 0.0112 & 0.0096 & 0.0100 & 0.0103 & 0.0212    \\ \hline
 RMSE-f(150) &  0.2690 & 0.1827 & 0.1313  & 0.0501 & 0.0607 & 0.0514 & 0.0373 & 0.0382 & 0.0763     \\ \hline
Log-likelihood ($\times 10^4$) & -14.4& 6.03 & 6.63& 6.59 & 6.61 & 6.60 & 6.61 & 6.63 & 6.40\\ \hline
\ctable{Training time}{(per epoch)}&  11.70s  & 11.78s & 11.79s & 11.98s & 11.98s & 12.10s & 12.44s & 13.40s & 26.75s   \\ \hline
Test time&  3.36s & 3.80s & 4.13s & 4.14s & 4.08s & 4.21s & 4.53s & 4.90s & 11.54s   \\ \hline
\end{tabular}}
%\caption{Comparison of log-likelihood, training time, and test time.}
\caption{Performance metrics for ROAD-EnKF at convergence with full observation ($d_u=d_y=256$) and different latent space dimension $d_z$. (Burgers example, \cref{ssec:Burgers}.)}\label{tb:burgers_table_2}
\end{table}
%\caption{Performance metrics for ROAD-EnKF at convergence %with full observation ($d_u=d_y=256$) and different %latent space dimension $d_z$. (Burgers example, %\cref{ssec:Burgers}.)}\label{tb:burgers_table_2}
%\end{figure}

\subsection{Kuramoto-Sivashinsky Equation}\label{ssec:KS}
In this subsection, we consider the Kuramoto-Sivashinsky (KS) equation for $u(x, s)$, where $u$ is a function of the spatial variable $x\in [0,\mathsf{L}]$ and continuous-time variable $s>0$:
\begin{equation}\label{eq:KS}
\begin{split}
    &\frac{\partial u}{\partial s} = -\nu \frac{\partial^4 u}{\partial x^4} - \frac{\partial^2 u}{\partial x^2} - u \frac{\partial u}{\partial x},\\
    &u(0, s) = u(\mathsf{L}, s) = 0,\\
    &\frac{\partial u}{\partial x}(0, s) = \frac{\partial u}{\partial x}(\mathsf{L},s) = 0,\\
    &u(x, 0) = u_0(x),
\end{split}
\end{equation}
Here $\nu$ is the viscosity parameter, and we set $\nu=0.05$, $\mathsf{L}=2$. We impose Dirichlet and Neumann boundary conditions to ensure ergodicity of the system \cite{blonigan2014least}. The KS equation was originally introduced by Kuramoto and Sivashinsky to model turbulence of reaction-diffusion systems \cite{kuramoto1976persistent} and propagation of flame \cite{sivashinsky1977nonlinear}.  Equation \cref{eq:KS} is discretized on $[0,\mathsf{L}]$ with equally-spaced grid points $0 = x_1 < x_2 < \cdots < x_M = \mathsf{L}$, using a second-order finite difference method. Setting $\Delta x\triangleq x_i - x_{i-1} = \frac{\mathsf{L}}{M-1}$,  we obtain the following ODE system:
\begin{equation}\label{eq:KS_discretized}
\begin{split}
&\frac{\partial u^{(i)}}{\partial s} = -\nu \frac{u^{(i-2)}-4u^{(i-1)}+6u^{(i)}-4u^{(i+2)}+u^{(i+2)}}{\Delta x^4} - \frac{u^{(i+1)}-2u^{(i)}+u^{(i-1)}}{\Delta x^2} - \frac{\big(u^{(i+1)}\big)^2-\big(u^{(i-1)}\big)^2}{4\Delta x},\\
&\qquad\qquad\qquad\qquad\qquad\qquad\qquad\qquad\qquad\qquad\qquad\qquad\qquad\qquad\qquad\qquad\qquad\qquad i=2,\dots, d_u-1,\\
&u^{(1)}(s) = u^{(d_u)}(s) = 0,\\
&u^{(0)}(s)=u^{(2)}(s), u^{(d_u+1)}(s)=u^{(d_u-1)}(s),\\
&u^{(i)}(0) = u_0(i\Delta x).
\end{split}
\end{equation}
The discretization method follows \cite{zhong2017reduced}. Here $u^{(i)}(s)$ is an approximation of $u(i\Delta x, s)$, the value of $u$ at the $i$-th spatial node and time $s$. Two ghost nodes $u^{(0)}$ and $u^{(d_u+1)}$ are added to account for Neumann boundary conditions, and are not regarded as part of the state. Equation \Cref{eq:KS_discretized} defines a flow map $F: u(s) \mapsto u(s + \Delta_s)$ for state variable $u$ with $d_u = M$, which we refer to as the true state dynamics model. We assume there is no noise in the dynamics, i.e., $Q=0$.

Similar to \cref{ssec:L63}, we consider two cases: full observation with $d_u=d_y=256$ and partial observation with $d_u=256$, $d_y=128$ (i.e., $c=1/2$). The initial conditions $u_0$ are generated at random from the attractor of the dynamical system, by simulating a long run beforehand. We generate $N_\train=512$ training data and $N_\test=20$ test data with the true state dynamics model defined through \cref{eq:KS_discretized} with $R_t=I_{d_y}$. We set the number of observations $T=450$ with time between observations $\Delta_s=0.1$. We set the forecast lead time $T_f=50$. The flow map $F$ is integrated using the fourth-order Runge–Kutta method with a fine step size $\Delta_s/10000$. The surrogate latent dynamics map $g_\alpha$ is parameterized as a two-layer fully connected NN, and is integrated using a fourth-order Runge-Kutta method with step size $\Delta_s^\text{int}=0.05$. The error covariance matrix $S_\beta$ in the latent dynamics is parametrized using a diagonal matrix with positive diagonal elements $\beta\in \R^{d_z}$. The decoder $\Dc_\gamma$ is parameterized as an FND, discussed in \cref{ssec:decoder_design}. Details of the network hyperparameters are listed in \cref{tb:exp_network_param}. The latent space dimension for ROAD-EnKF is set to $d_z=40$. The ensemble size for both AD-EnKF and ROAD-EnKF is set to $N=100$.

\begin{figure}[!htb]
\centering
\begin{subfigure}[b]{\textwidth}
\includegraphics[width=\textwidth]{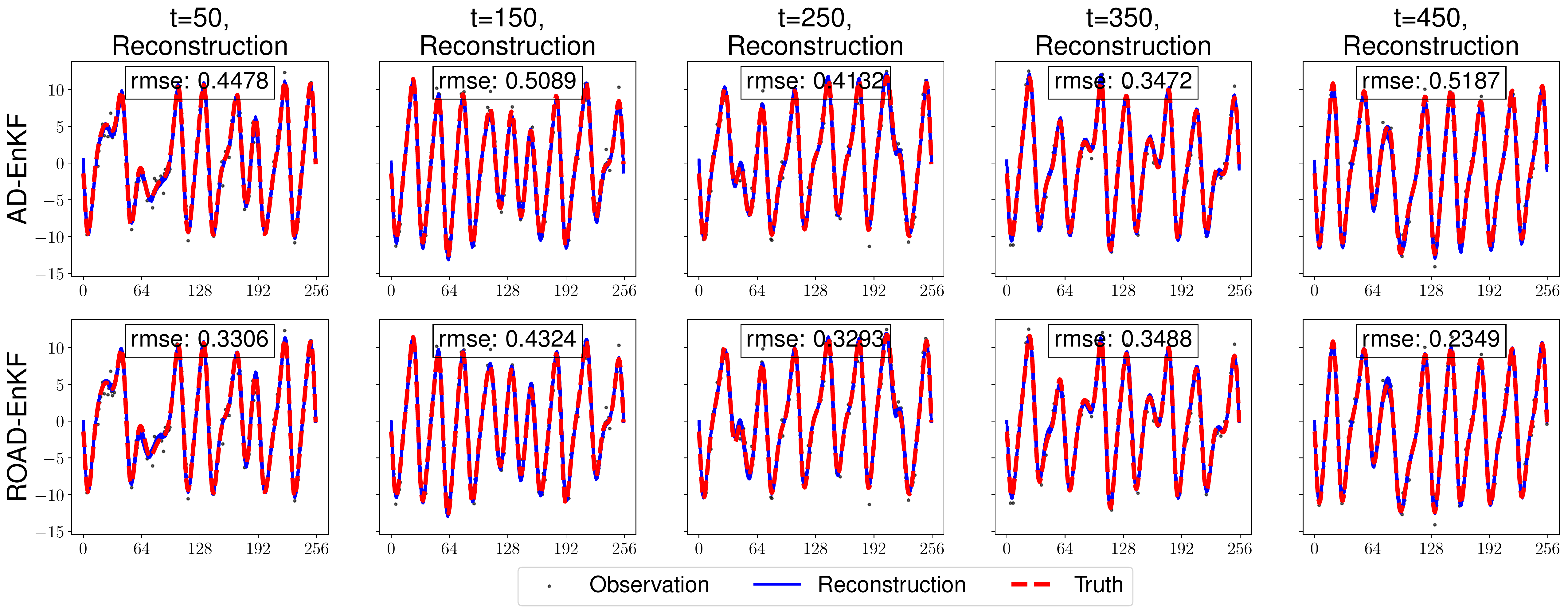}
\end{subfigure}
\begin{subfigure}[b]{\textwidth}
\includegraphics[width=\textwidth]{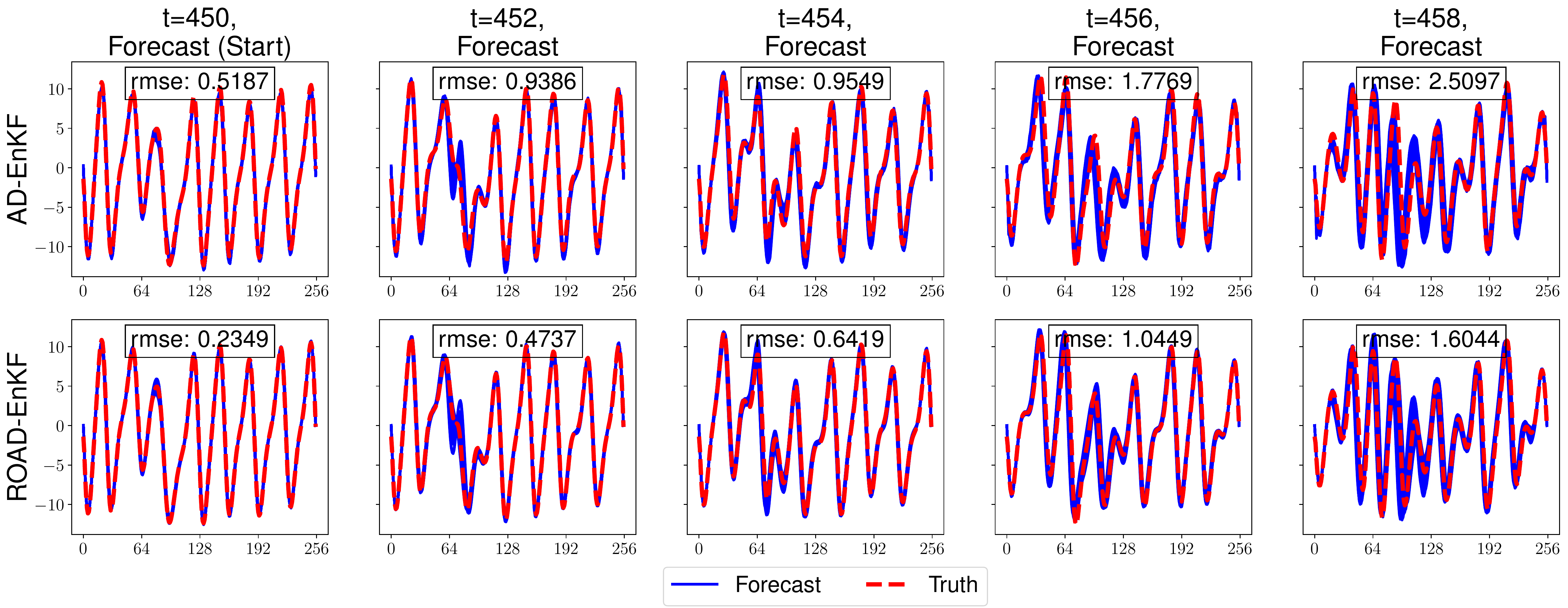}
\end{subfigure}
\caption{State reconstruction (upper half) and forecast (lower half) performance with partial observation ($d_u=256$, $d_y=128$) on the KS example in \cref{ssec:KS}. For each method,  the reconstructed states $u_t$ (blue) for a single test sequence are plotted for $t=50, 150, 250, 350, 450$ (column), and the forecasted states (blue) for a single test sequence are plotted for $t=450$ (start of forecast), $452, 454, 456, 458$ (column). The true values of the 256-dimensional states are plotted in red dashed lines, along with the noisy observations in black dots. Both AD-EnKF and ROAD-EnKF perform probabilistic state reconstructions and forecast through particles (all plotted in blue). The reconstruction/forecast RMSEs are computed for each plot. }\label{fig:ks_recon_partial}
\end{figure}

\begin{figure}[!htb]
\centering
\begin{subfigure}[b]{0.55\textwidth}
   \includegraphics[width=1\linewidth]{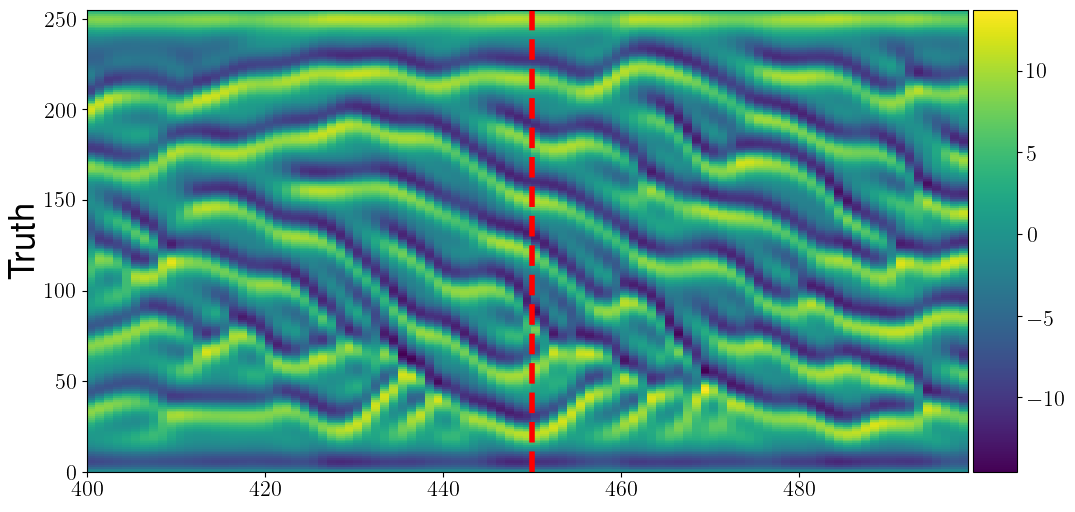}
   \caption{Ground truth.}
\end{subfigure}

\begin{subfigure}[b]{\textwidth}
   \includegraphics[width=1\linewidth]{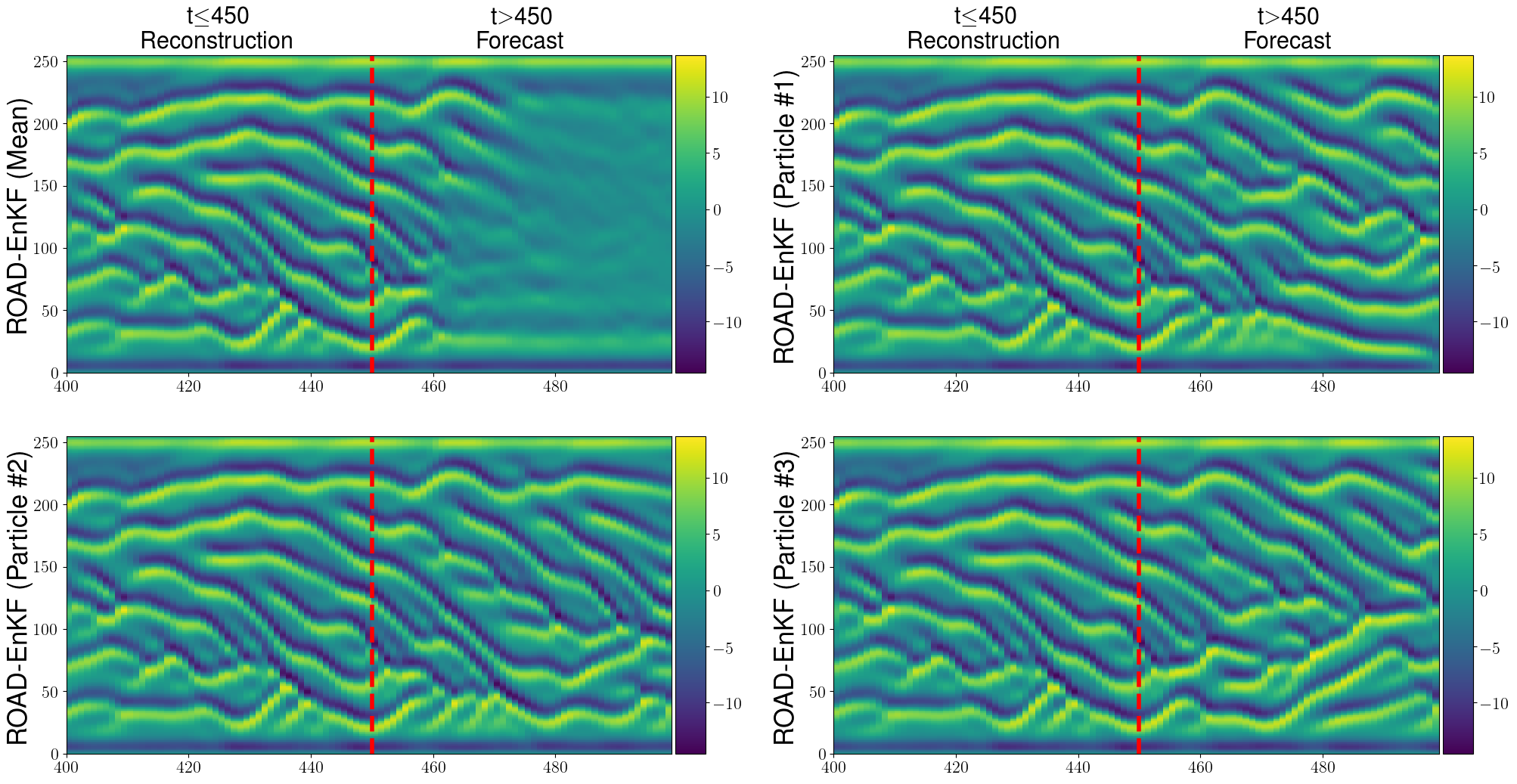}
   \caption{Reconstruction and forecast.}
\end{subfigure}
\caption{Contour plot of state reconstruction and forecast output of ROAD-EnKF with partial observation ($d_u=256$, $d_y=128$) on the KS example in \cref{ssec:KS}, as well as the ground truth (top). The particle means of reconstructed and forecasted states for a single test sequence are plotted, for each state dimension (y-axis) and time (x-axis). The reconstructed and forecasted states of three randomly chosen particles are also plotted individually.
} 
\label{fig:ks_contour_partial} 
\end{figure}

In \cref{tb:ks_table} we list the performance metrics of AD-EnKF and ROAD-EnKF with full and partial observation. SINDy-AE is not listed here as we find it unable to capture the dynamics for any choice of latent space dimension. The state reconstruction and forecast performance on a single instance of test data are plotted in \cref{fig:ks_recon_partial} (snapshots), and \cref{fig:ks_contour_partial} (contour plot, ROAD-EnKF) for the partial observation case. Corresponding plots with full observation are shown in Figures \ref{fig:ks_recon} and \ref{fig:ks_contour} in the appendix. We find that ROAD-EnKF is able to reconstruct the states with lower RMSE than AD-EnKF in both full observation and partial observation cases. Both methods can produce meaningful forecast multiple steps forward into the future. ROAD-EnKF achieves a higher forecast RMSE than AD-EnKF in full observation case, while having a lower forecast RMSE in partial observation case. Although ROAD-EnKF does not consistently have a better forecast performance than AD-EnKF due to the difficulty of finding a reduced-order representation for the highly chaotic system, we find that its performance is not much impacted by partial observation. Moreover, it is two times more efficient than AD-EnKF in both training and testing, due to the times saved for simulating a cheaper surrogate model and running the EnKF algorithm in a lower dimensional space.  Notice in \cref{fig:ks_contour_partial}(b) and \cref{fig:ks_contour}(b) that, although the predictive means of all particles are `smoothed' when passing a certain time threshold, each particle individually produces nontrivial forecasts for a larger number of time steps into the future, thus illustrating the variability of particle forecasts and the stochastic nature of state reconstruction and forecast in our ROAD-EnKF framework.

% \begin{figure}[!htb]
% \begin{subfigure}[c]{\linewidth}
% \centering
% \includegraphics[height=5cm]{KS/new_bar.pdf}
% \subcaption{Comparison of RMSE-r, RMSE-f(1), and RMSE-f(5).}
% \vspace{0.5cm}
% \end{subfigure}
% \begin{subtable}{\textwidth}
\begin{table}[!htb]
\centering
\begin{tabular}{|c|c|c?{0.5mm}c|c|}
\hline
            & \ctable{AD-EnKF}{(full)} & \ctable{ROAD-EnKF}{(full)} & \ctable{AD-EnKF}{(partial)} & \ctable{ROAD-EnKF}{(partial)}  \\ \hline
RMSE-r     & 0.4658 & 0.3552 & 0.4686 &0.3589    \\ \hline
RMSE-f(1)  & 0.5137  & 0.5626  & 0.6231 &0.5644   \\ \hline
RMSE-f(5)  & 1.0910 & 1.2734 & 1.4669 & 1.3780   \\ \hline
Log-likelihood & $-1.89\times 10^6$ & $-1.88\times 10^6$ & $-9.33 \times 10^5$ & $-9.07 \times 10^5$   \\ \hline
Training time (per epoch)&  28.92s  & 12.53s & 28.72s & 12.61s    \\ \hline
Test time & 12.35s  & 5.11s & 6.22s & 4.39s   \\ \hline
\end{tabular}
% \subcaption{Comparison of log-likelihood, training time, and test time. Log-likelihood evaluation is not supported for SINDy-AE.}
% \end{subtable}
\caption{Performance metrics for different algorithms at convergence. (KS example, \cref{ssec:KS}.)}\label{tb:ks_table}
\end{table}

% \begin{table}[!htb]
% \centering
% \begin{tabular}{|c|c|c?{0.5mm}c|c|}
% \hline
%             & \ctable{AD-EnKF}{(full)} & \ctable{ROAD-EnKF}{(full)} & \ctable{AD-EnKF}{(partial)} & \ctable{ROAD-EnKF}{(partial)}  \\ \hline
% RMSE-r     & 0.4658 & 0.3552 & 0.4686 &0.3589    \\ \hline
% RMSE-f(1)  & 0.5137  & 0.5626  & 0.6231 &0.5644   \\ \hline
% RMSE-f(5)  & 1.0910 & 1.2734 & 1.4669 & 1.3780   \\ \hline
% Log-likelihood & $-1.89\times 10^6$ & $-1.88\times 10^6$ & $-9.33 \times 10^5$ & $-9.07 \times 10^5$   \\ \hline
% Training time (per epoch)&  28.92s  & 12.53s & 28.72s & 12.61s    \\ \hline
% Test time & 12.35s  & 5.11s & 6.22s & 4.39s   \\ \hline
% \end{tabular}
% \caption{Performance metrics for different algorithms at convergence. (KS example, \cref{ssec:KS}.)}
% \label{tb:ks_table}
% \end{table}

\section{Conclusions and Future Directions}\label{sec:conclusions}
This paper introduced a computational framework to reconstruct and forecast a partially observed state that evolves according to an unknown or expensive-to-simulate dynamical system. Our ROAD-EnKFs use an EnKF algorithm to estimate by maximum likelihood a surrogate model for the dynamics in a latent space, as well as a decoder from latent space to state space.
Our numerical experiments demonstrate the computational advantage of co-learning an inexpensive surrogate model in latent space together with a decoder, rather than a more expensive-to-simulate dynamics in state space. 

The proposed computational framework accommodates partial observation of the state, does not require time derivative data, and enables uncertainty quantification. In addition, it provides significant algorithmic flexibility through the choice of latent space, surrogate model for the latent dynamics, and decoder design. In this work, we showed that accurate and cheap reconstructions and forecasts can be obtained by choosing an inexpensive NN surrogate model, and a decoder inspired by recent ideas from operator learning. While adequate choice of NN architecture and decoder may be problem-specific, an important question for further research is to derive guidelines and physics-informed NNs that are  well-suited for certain classes of problems.

\section*{Acknowledgments}
The authors are grateful to Melissa Adrian for her generous feedback on an earlier version of this manuscript.
YC was partially supported by NSF DMS-2027056 and NSF OAC-1934637.
DSA is grateful for the support of NSF DMS-2237628, NSF DMS-2027056, DOE DE-SC0022232, and the BBVA Foundation.
RW is grateful for the support of DOD FA9550-18-1-0166, DOE DE-AC02-06CH11357, NSF OAC-1934637, NSF DMS-1930049, and NSF DMS-2023109. 

\bibliographystyle{siamplain}
\bibliography{references}

\appendix

\section{Improving AD-EnKF with Spectral Convolutional Layers}\label{sec:improving-AD-EnKF}
This appendix discusses an enhancement of the AD-EnKF algorithm \cite{chen2021auto}, used for numerical comparisons in Section \ref{sec:numerical}. AD-EnKF runs EnKF on the full-order SSM \eqref{eq:ref_model_state}-\eqref{eq:ref_model_init} and learns the parameter $\theta = (\alpha^\top,\beta^\top)^\top$ by auto-differentiating through a similarly defined log-likelihood objective, as in \cref{ssec:enkf_loglike}. A high dimension of $u$ makes challenging the NN parameterization of $F_\alpha$ (resp. $f_\alpha$ in the ODE case) in the state dynamics model \cref{eq:ref_model_state}. In particular, the local convolutional NN used in \cite{chen2021auto} does not perform well in the high-dimensional numerical experiments considered in \cref{sec:numerical}. We thus propose a more flexible NN parameterization of $F_\alpha$ (resp. $f_\alpha$) using the idea of spectral convolutional layers.

We design $F_\alpha$ (resp. $f_\alpha$) in a way similar to the Fourier Neural Decoder, but without the complex linear layer and IDFT step at the beginning. That is, we start with a state variable $u' \in \R^{d_u}$ as the input, iteratively apply \cref{eq:spec_layer} with $v_0=u'$ to get $v_L\in \R^{n_L \times d_u}$, followed by a fully-connected network applied over the channel dimension to get the output $u\in \R^{d_u}$. The architecture is the same as \cref{fig:network}(a) but we start at $v_0$ instead of $z$.

\section{Additional Materials: Burgers Example}\label{sec:addfig-appen-burgers}
For SINDy-AE, we use a finite difference approximation computed from data $y_{1:T}$ to approximate the exact time-derivative. The latent space dimension for SINDy-AE is set to 6. Increasing it does not further enhance the performance, but increases the computational cost.

%\begin{figure}[!htb]
%\begin{subfigure}[c]{\linewidth}
%\centering
%\includegraphics[height=5cm]{Burgers/new_bar_3.pdf}
%\subcaption{Comparison of RMSE-r, RMSE-f(30), and RMSE-f(150).}
%\vspace{0.5cm}
%\end{subfigure}
\begin{table}[!htb]
\centering
{\footnotesize
\begin{tabular}{|c|c|c|c|c|c|c|}
\hline
            & \ctable{AD-EnKF}{(FC, Euler)}& \ctable{AD-EnKF}{(FC, RK4)}& \ctable{AD-EnKF}{(Fourier, Euler)} & \ctable{AD-EnKF}{(Fourier, RK4)} & \ctable{ROAD-EnKF}{(FC, Euler)} & \ctable{ROAD-EnKF}{(FC, RK4)}  \\ \hline
 RMSE-r     &0.1023 & 0.0934 & 0.0537 & 0.0102 & 0.0045 &0.0044    \\ \hline
 RMSE-f(30) &0.0999 & 0.0831 & 0.1302 & 0.0212  & 0.0100  &0.0096   \\ \hline
 RMSE-f(150) &0.1971 & 0.1608 & 0.2908 & 0.0763 & 0.0664 & 0.0514   \\ \hline
Log-likelihood & $2.31\times 10^4$ & $2.87\times 10^4$ & $5.41\times 10^4$ & $6.40 \times 10^4$ & $6.60\times 10^4$ & $6.57 \times 10^4$   \\ \hline
Training time (per epoch) &4.97s & 5.80s &  12.31s & 26.75s  & 11.21s & 12.10s    \\ \hline
Test time &2.29s & 2.97s & 4.79s & 11.54s  & 3.28s & 4.21s   \\ \hline
\end{tabular}}
\caption{Ablation study: AD-EnKF versus ROAD-EnKF with different NN parameterization and numerical integration methods for surrogate dynamics (FC: NN with fully-connected layers; Fourier: NN with Fourier layers; Euler: Euler method for ODE integration; RK4: fourth-order Runge Kutta method for ODE integration). Switching from RK4 to Euler method while keeping the same NN configuration gives a computational speed-up, and the speed-up is more noticeable when the NN involves Fourier layers. However, after the switch, the accuracy drops more significantly for AD-EnKF than for ROAD-EnKF. The best configuration for AD-EnKF (Fourier with RK4) still yields a lower accuracy compared to both ROAD-EnKF configurations, while taking more time to compute. (Burgers example, full observation case, \cref{ssec:Burgers}.)}\label{tb:burgers_euler_table}
\end{table}
% \caption{Ablation study: AD-EnKF versus ROAD-EnKF with different NN parameterization and numerical integration methods for surrogate dynamics (FC: NN with fully-connected layers; Fourier: NN with Fourier layers; Euler: Euler method for ODE integration; RK4: fourth-order Runge Kutta method for ODE integration). Switching from RK4 to Euler method while keeping the same NN configuration gives a computational speed-up, and the speed-up is more noticeable when the NN involves Fourier layers. However, after the switch, the accuracy drops more significantly for AD-EnKF than for ROAD-EnKF. The best configuration for AD-EnKF (Fourier with RK4) still yields a lower accuracy compared to both ROAD-EnKF configurations, while taking more time to compute. (Burgers example, full observation case, \cref{ssec:Burgers}.)}\label{tb:burgers_euler_table}
%\end{figure}

\begin{figure}[H]
\centering
\begin{subfigure}[b]{\textwidth}
\includegraphics[width=\textwidth]{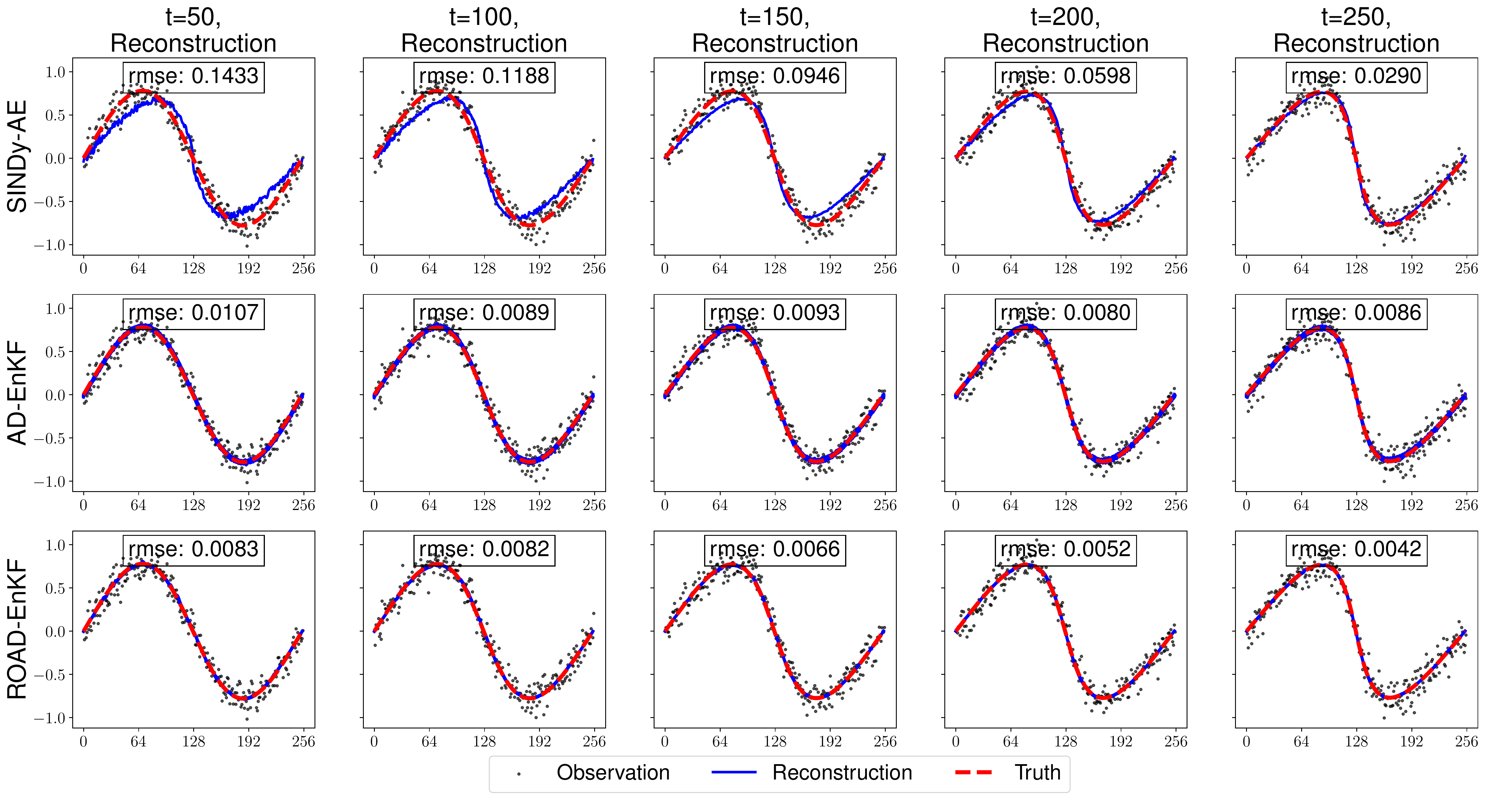}
\end{subfigure}
\begin{subfigure}[b]{\textwidth}
\includegraphics[width=\textwidth]{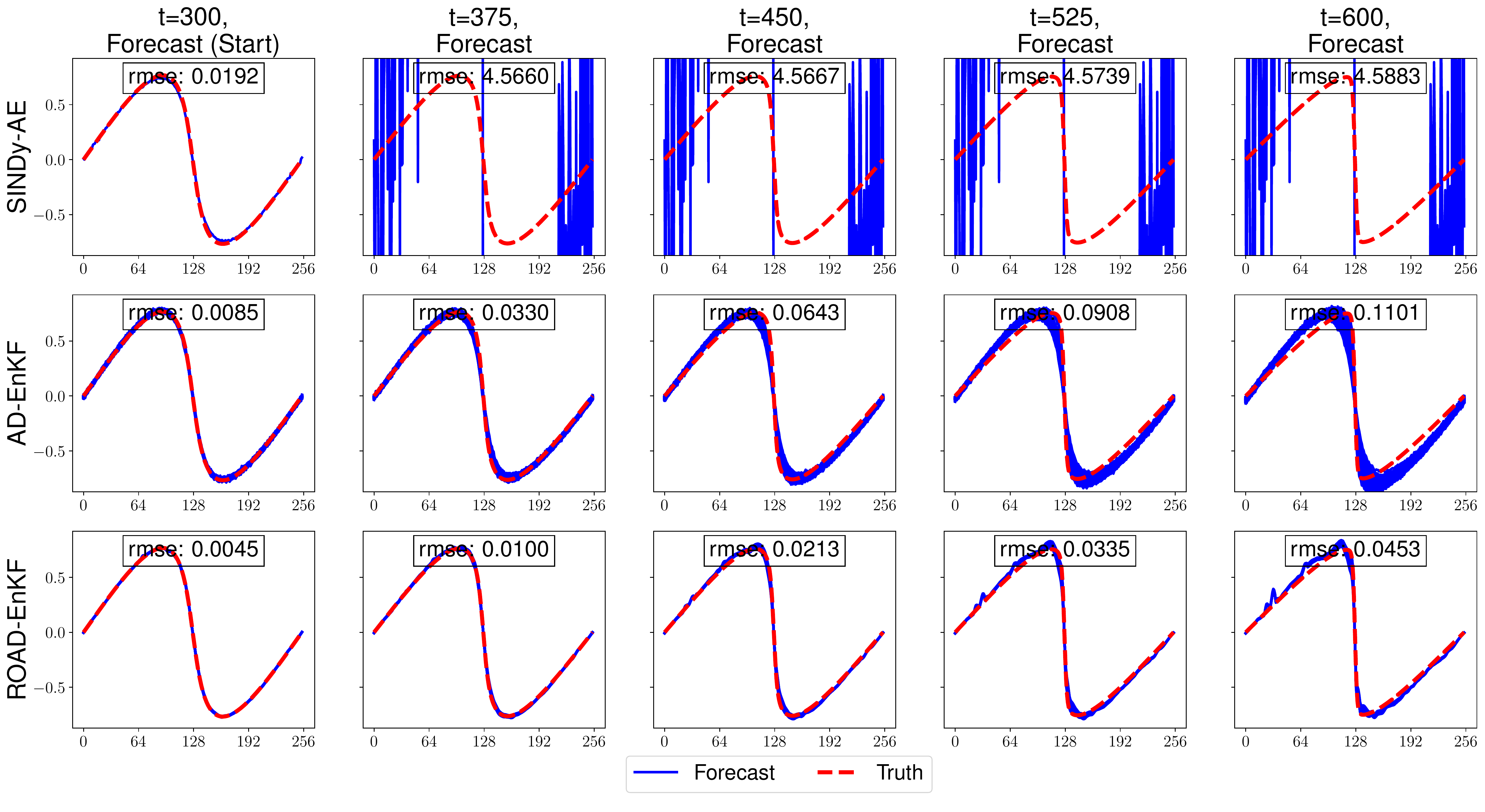}
\end{subfigure}
\caption{State reconstruction (upper half) and forecast (lower half) performance with full observation ($d_u=d_y=256$) on the Burgers example in \cref{ssec:Burgers}. For each method, the reconstructed states $u_t$ (blue) for a single test sequence are plotted for $t=50, 100, 150, 200, 250$ (column), and the forecasted states (blue) for a single test sequence are plotted for $t=300$ (start of forecast), $375, 450, 525, 600$ (column). The true values of the 256-dimensional states are plotted in red dashed lines, along with the noisy observations in black dots. Both AD-EnKF and ROAD-EnKF perform probabilistic state reconstructions and forecast through particles (all plotted in blue), while SINDy-AE only provides point estimates. The reconstruction/forecast RMSEs are computed for each plot.  }\label{fig:burgers_recon}
\end{figure}

\begin{figure}[!htb]
\centering
\begin{subfigure}[b]{0.55\textwidth}
   \includegraphics[width=1\linewidth]{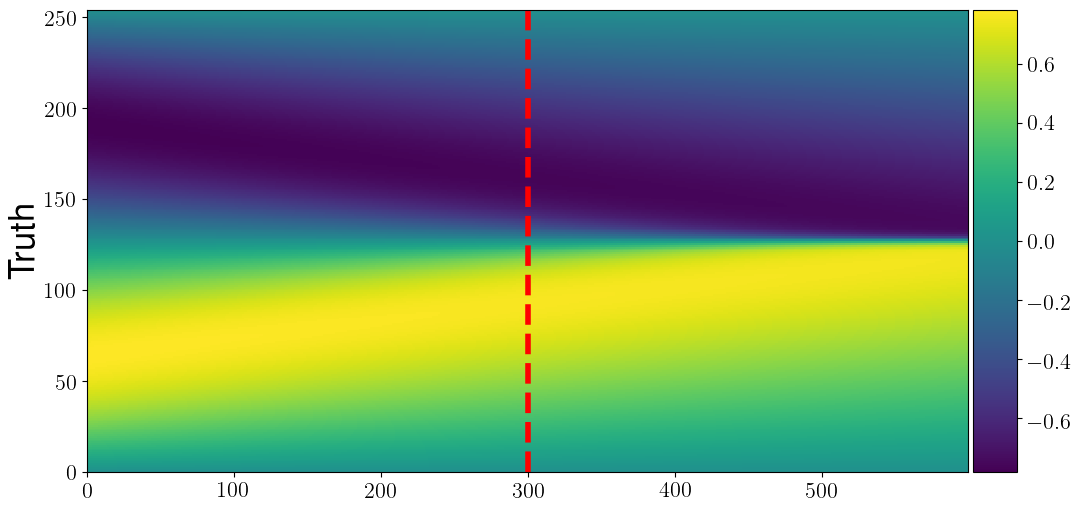}
   \caption{Ground truth.}
\end{subfigure}

\begin{subfigure}[b]{\textwidth}
   \includegraphics[width=1\linewidth]{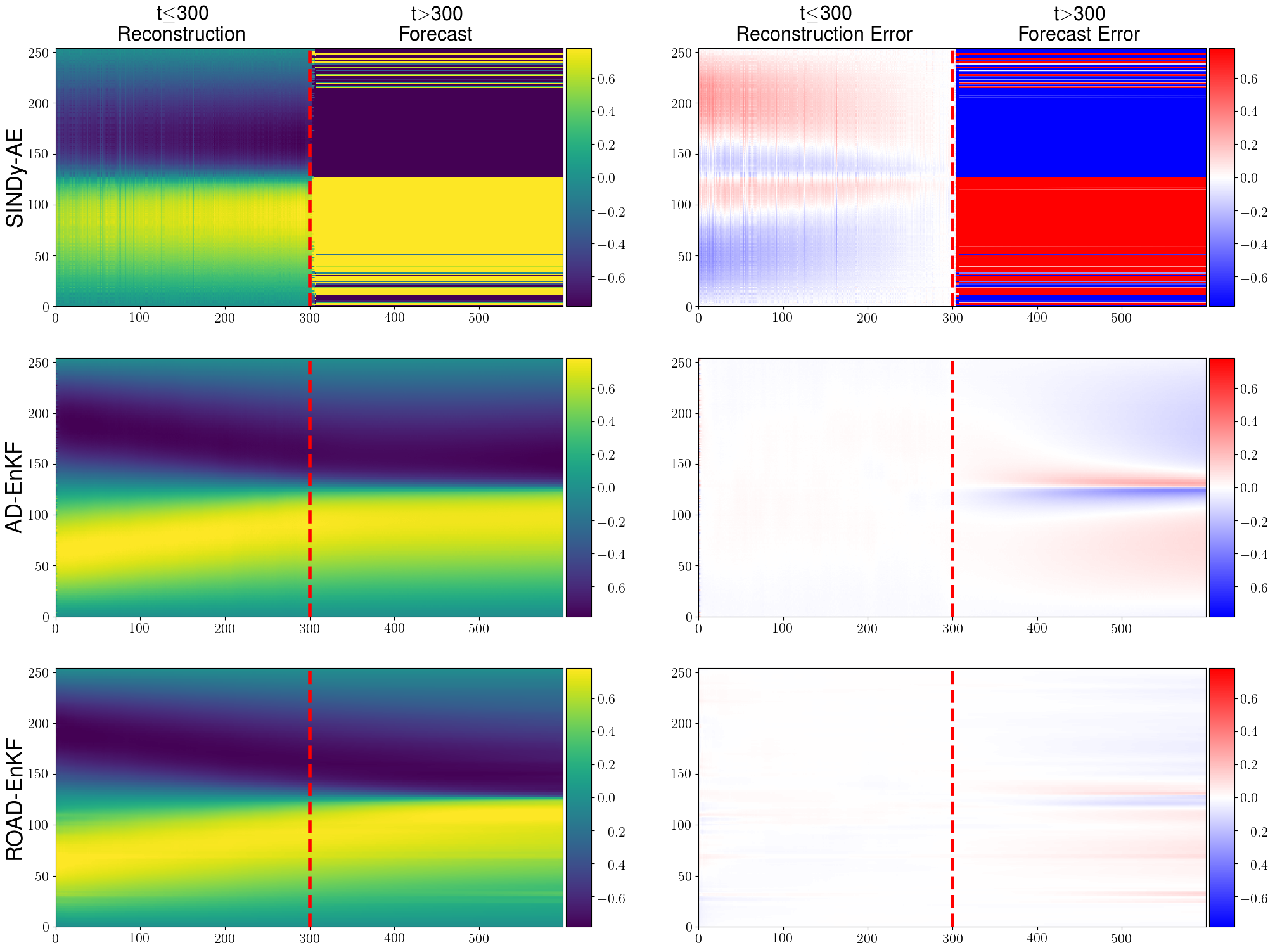}
   \caption{Reconstruction and forecast.}

\end{subfigure}
\caption{Contour plot of state reconstruction and forecast output with full observation ($d_u=d_y=256$) on the Burgers example in \cref{ssec:Burgers}, as well as the ground truth (top). For each method (row), the reconstructed and forecasted states (left column) for a single test sequence are plotted, for each state dimension (y-axis) and time (x-axis). The error compared to the ground truth are plotted in the right column. For both AD-EnKF and ROAD-EnKF we use particle means as point estimates.}
\label{fig:burgers_contour} 
\end{figure}

% \includegraphics[width=\textwidth]{Burgers/contour.png}
% \caption{}\label{fig:burgers_contour}
% \end{figure}

\section{Additional Materials: Kuramoto-Sivashinky Example}\label{sec:addfig-appen-ks}
\begin{figure}[!htb]
\centering
\begin{subfigure}[b]{\textwidth}
\includegraphics[width=\textwidth]{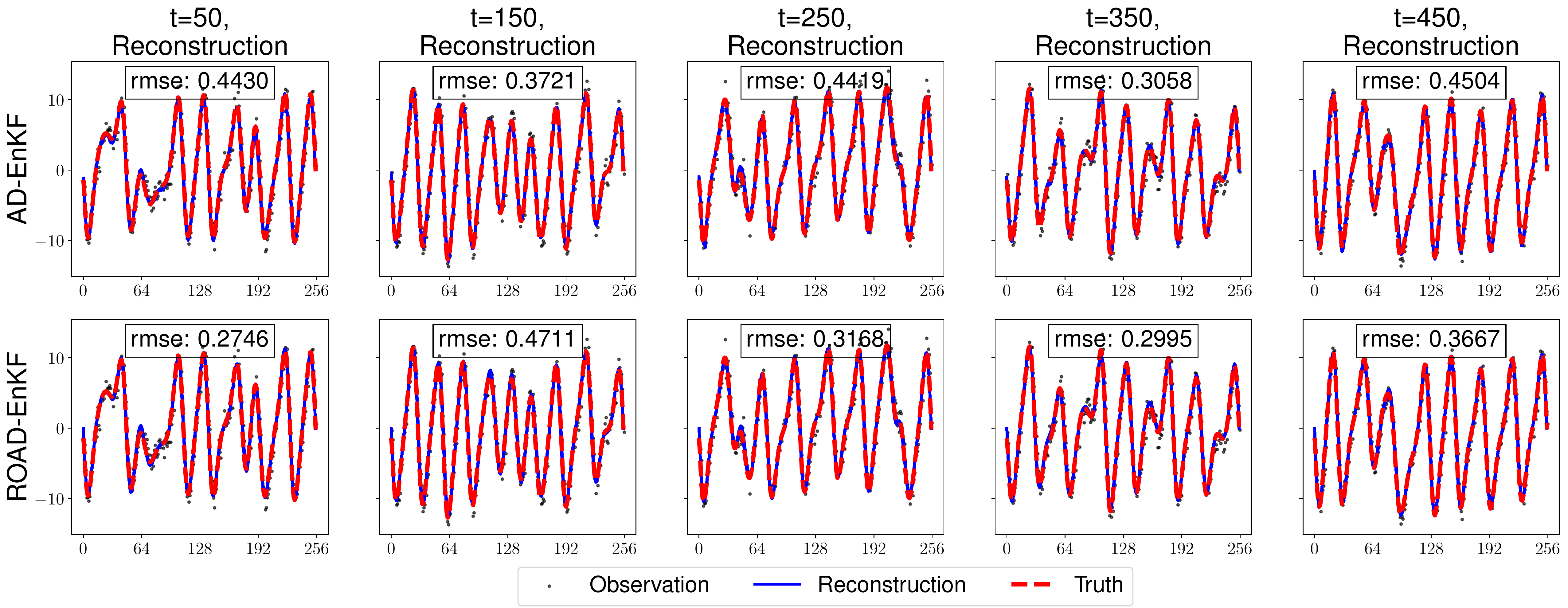}
\end{subfigure}
\begin{subfigure}[b]{\textwidth}
\includegraphics[width=\textwidth]{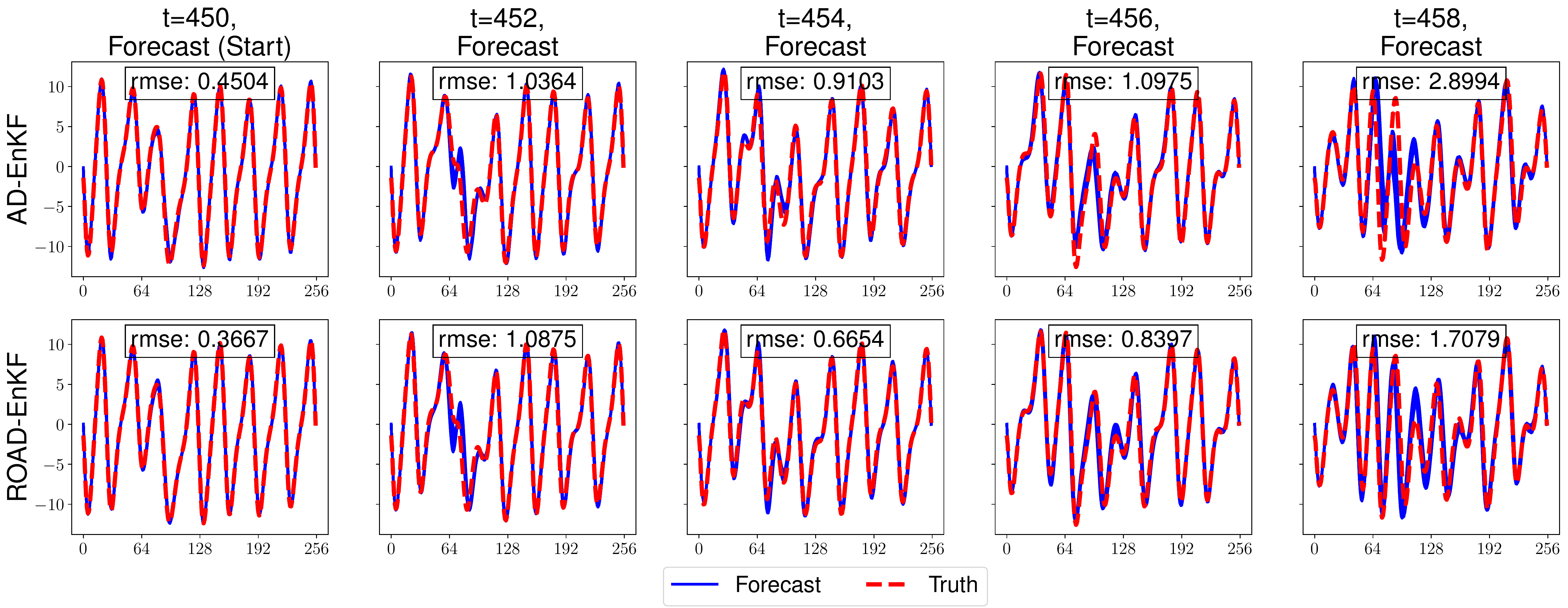}
\end{subfigure}
\caption{State reconstruction (upper half) and forecast (lower half) performance with full observation ($d_u=d_y=256$) on the KS example in \cref{ssec:KS}. For each method,  the reconstructed states $u_t$ (blue) are plotted for $t=50, 150, 250, 350, 450$ (column), and the forecasted states (blue) are plotted for $t=450$ (start of forecast), $452, 454, 456, 458$ (column). The true values of the 256-dimensional states are plotted in red dashed lines, along with the noisy observations in black dots. Both AD-EnKF and ROAD-EnKF perform probabilistic state reconstructions and forecast through particles (all plotted in blue). The reconstruction/forecast RMSEs are computed for each plot. }\label{fig:ks_recon}
\end{figure}

\begin{figure}[!htb]
\centering
\begin{subfigure}[b]{0.55\textwidth}
   \includegraphics[width=1\linewidth]{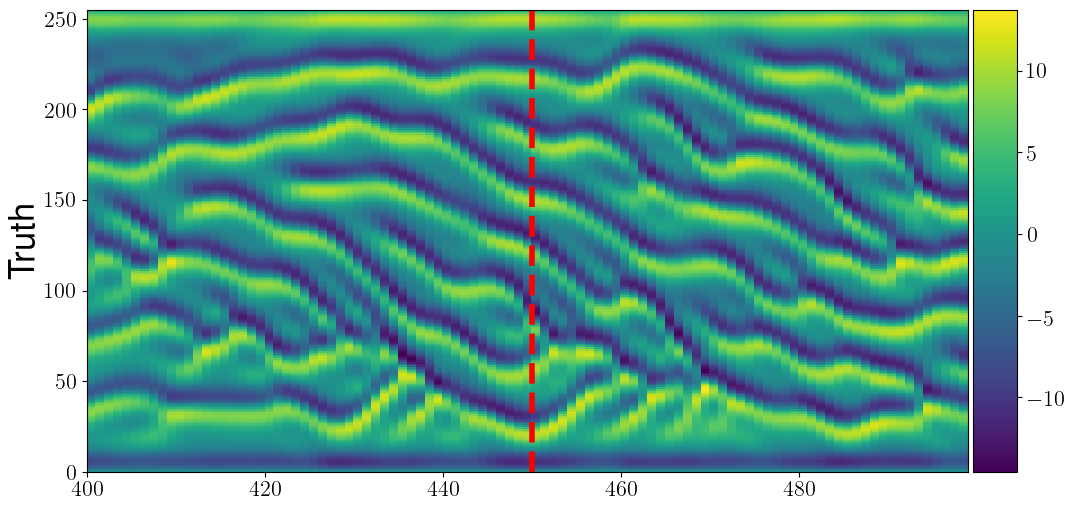}
   \caption{Ground truth.}
\end{subfigure}

\begin{subfigure}[b]{\textwidth}
   \includegraphics[width=1\linewidth]{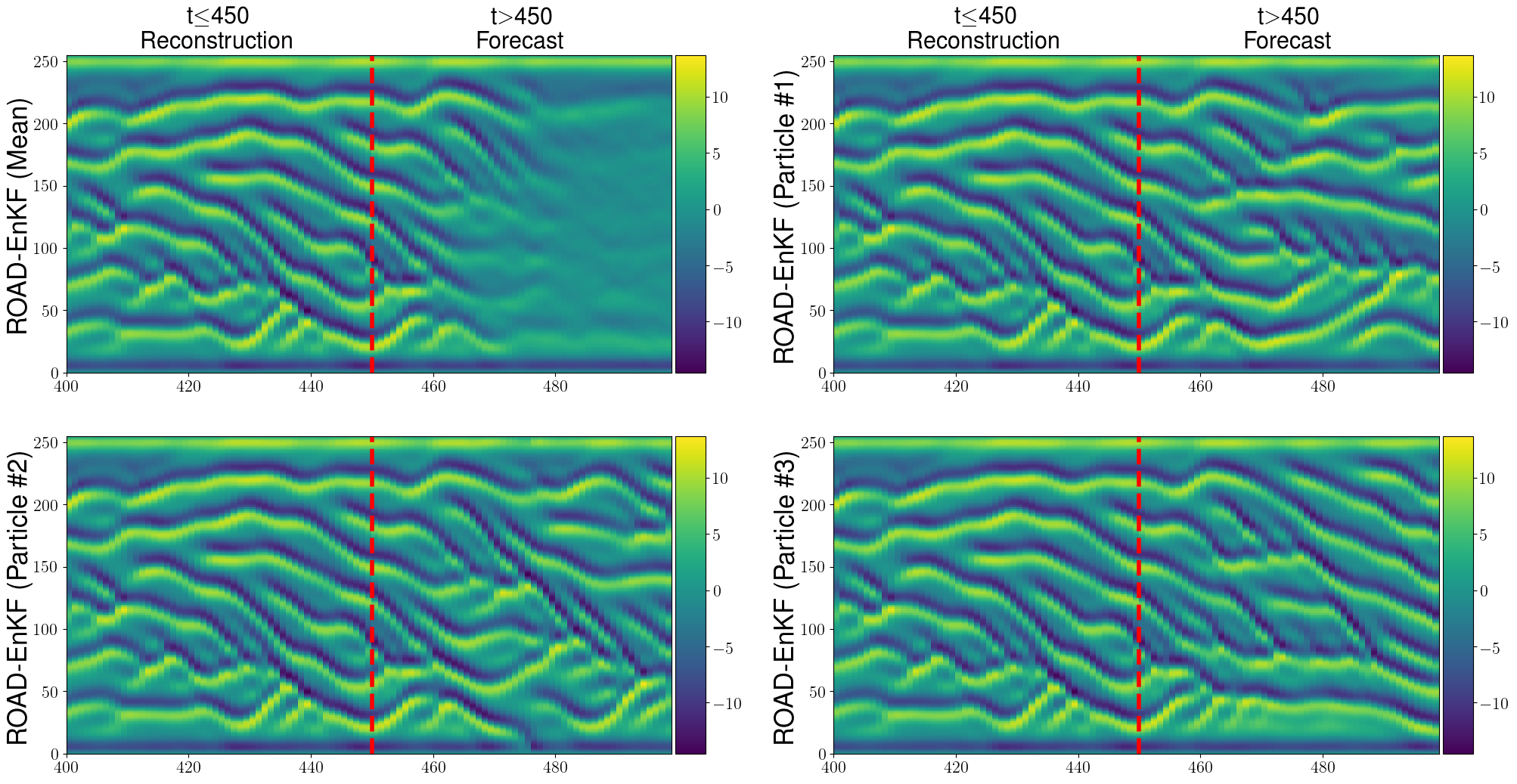}
   \caption{Reconstruction and forecast.}
\end{subfigure}
\caption{Contour plot of state reconstruction and forecast output of ROAD-EnKF with full observation ($d_u=d_y=256$) on the KS example in \cref{ssec:KS}, as well as the ground truth (top). The particle means of reconstructed and forecasted states are plotted, for each state dimension (y-axis) and time (x-axis). The individual reconstructed and forecasted states of three randomly chosen particles are also plotted.}
\label{fig:ks_contour} 
\end{figure}

\end{document}

%% file: shared.tex
\usepackage{sidecap}
\usepackage{enumitem}
\usepackage{lipsum}
\usepackage{graphicx}
\usepackage{subcaption}
\usepackage{epstopdf}
\ifpdf
  \DeclareGraphicsExtensions{.eps,.pdf,.png,.jpg}
\else
  \DeclareGraphicsExtensions{.eps}
\fi

\usepackage[capitalize]{cleveref}
\crefname{section}{Section}{Section}
\crefname{subsection}{Subsection}{Subsection}
\crefname{subsubsection}{Subsection}{Subsection}
\crefname{equation}{}{equation}
\crefname{figure}{Figure}{Figure}
\crefname{table}{Table}{Table}
\crefname{algorithm}{Algorithm}{Algs.}
\crefformat{section}{Section #2#1#3}
\crefmultiformat{section}{\S\S#2#1#3}{ and~#2#1#3}{, #2#1#3}{, and~#2#1#3}
\AtBeginDocument{%
\crefformat{subsection}{Subsection #2#1#3}
\Crefformat{subsection}{Subsection #2#1#3}
\crefmultiformat{subsection}{Subsection #2#1#3}{ and~#2#1#3}{, #2#1#3}{, and~#2#1#3}
\crefformat{subsubsection}{\S#2#1#3}
\Crefformat{subsubsection}{\S#2#1#3}
}
\crefname{hypothesis}{Hypothesis}{Hypotheses}

\usepackage{amsopn}

\makeatletter
\newcommand*{\addFileDependency}[1]{% argument=file name and extension
  \typeout{(#1)}% latexmk will find this if $recorder=0 (however, in that case, it will ignore #1 if it is a .aux or .pdf file etc and it exists! if it doesn't exist, it will appear in the list of dependents regardless)
  \@addtofilelist{#1}% if you want it to appear in \listfiles, not really necessary and latexmk doesn't use this
  \IfFileExists{#1}{}{\typeout{No file #1.}}% latexmk will find this message if #1 doesn't exist (yet)
}
\makeatother

\newcommand{\email}[1]{\protect\href{mailto:#1}{#1}}

\usepackage{bm} % boldmath must be called after the package

\usepackage{bbold}

\usepackage{amsmath,amsbsy,amsgen,amscd,amsfonts,amssymb} 

\usepackage{setspace}

\usepackage[font=small,labelfont=bf]{caption}

\usepackage{algorithm}
\usepackage{algpseudocode}
\usepackage{comment}

\usepackage{tikz}
\usetikzlibrary{fit,positioning,arrows,automata,calc}
\usepackage{mathtools}
\mathtoolsset{centercolon}  % Makes := typeset correctly for definitions

\renewcommand{\phi}{\varphi}

\renewcommand{\triangleq}{:=}

\providecommand{\mathbbm}{\mathbb} % In case we don't load bbm

\newcommand{\R}{\mathbbm{R}}
\newcommand{\C}{\mathbbm{C}}

\renewcommand{\L}{\mathcal{L}}

\definecolor{mygreen}{rgb}{0.1,0.75,0.2}

\newcommand{\Nc}{\mathcal{N}}

\usepackage{pifont}
\usepackage[percent]{overpic}
\usepackage{multirow}
\usepackage{float}
\usepackage{array}
\newcolumntype{?}[1]{!{\vrule width #1}}

\newcommand{\dd}{\text{d}}

\newcommand{\I}{\mathcal{I}}
\newcommand*{\B}[1]{\ifmmode\bm{#1}\else\textbf{#1}\fi}

\newcommand{\EnKF}{{\footnotesize \text{EnKF}}}

\newcommand{\train}{{\footnotesize \text{train}}}
\newcommand{\test}{{\footnotesize \text{test}}}

\newcommand{\LEnKF}{\L_{\EnKF}}

\newcommand{\ctable}[2]{\begin{tabular}[c]{@{}c@{}} #1\\ #2\end{tabular}}
\newcommand{\rmsef}{\text{RMSE-f}}

\newcommand{\rmser}{\text{RMSE-r}}

\newcommand{\tn}[1]{{#1}_t^n}
\newcommand{\tnf}[1]{ \widehat{#1}_t^{\, n}}

\newcommand{\tmn}[1]{{#1}_{t-1}^n}

\newcommand{\ton}[1]{{#1}_t^{1:N}}

\newcommand{\iidsim}{\overset{\text{i.i.d.}}{\sim}}
\newcommand{\nt}{\nabla_\theta}

\algnewcommand{\LineComment}[2]{\Statex \hskip #1 // \textit{#2}}
\newcommand{\DFT}{{\text{DFT}}}
\newcommand{\IDFT}{{\text{IDFT}}}
\newcommand{\logdet}{{\text{logdet}}}
\newcommand{\aug}{{\text{aug}}}
\newcommand{\cmark}{\ding{51}}%
\newcommand{\xmark}{\ding{55}}%

\newcommand{\Gc}{G}

\newcommand{\Sc}{S}
\newcommand{\Hc}{\mathcal{H}}
\newcommand{\Dc}{D}
\newcommand{\Kc}{K}

\usepackage[normalem]{ulem}

\usepackage{graphicx}

\makeatletter
\protected\def\tikz@nonactivecolon{\ifmmode\mathrel{\mathop\ordinarycolon}\else:\fi} 
\makeatother

\counterwithin{table}{section}
\counterwithin{algorithm}{section}